\documentclass[final]{cvpr}

\usepackage{times}
\usepackage{graphicx}
\usepackage{amsmath}
\usepackage{amssymb}
\usepackage{wasysym}
\usepackage{changepage}
\usepackage{ragged2e}
\usepackage{braket}
\usepackage{booktabs}
\usepackage{etoolbox}
\usepackage{multirow}
\usepackage{topcapt}
\usepackage{float}
\usepackage{appendix}
\usepackage{subfig}
\usepackage{lipsum}
\def\cvprPaperID{} 
\def\confYear{CVPR 2021}
\begin{document}

  \vspace{-10mm} 
\title{BGT-Net: Bidirectional GRU Transformer Network for Scene Graph Generation}
  \vspace{-10mm}

\author{
 Naina Dhingra\textsuperscript{}\space\space\space\space  \hspace{40mm} Florian Ritter\textsuperscript{}\space\space\space\space  \hspace{40mm} Andreas
 Kunz\textsuperscript{}\space\space\space\space 
  \vspace{6px} \\
\textsuperscript{}Innovation Center Virtual Reality, ETH Zurich\space\space\space\space \\
{\tt\small $\{$ndhingra, kunz$\}$@iwf.mavt.ethz.ch; ritterf@ethz.ch}}

  \vspace{-40mm}

\maketitle

\begin{abstract}
   Scene graphs are nodes and edges consisting of objects and object-object relationships, respectively. Scene graph generation (SGG) aims to identify the objects and their relationships. We propose a bidirectional GRU (BiGRU) transformer network (BGT-Net) for the scene graph generation for images. This model implements novel object-object communication to enhance the object information using a BiGRU layer. Thus, the information of all objects in the image is available for the other objects, which can be leveraged later in the object prediction step. This object information is used in a transformer encoder to predict the object class as well as to create object-specific edge information via the use of another transformer encoder. To handle the dataset bias induced by the long-tailed relationship distribution, softening with a log-softmax function and adding a bias adaptation term to regulate the bias for every relation prediction individually showed to be an effective approach. We conducted an elaborate study on experiments and ablations using open-source datasets, i.e., Visual Genome, Open-Images, and Visual Relationship Detection datasets, demonstrating the effectiveness of the proposed model over state of the art.

 \end{abstract}
    \vspace{-6mm}

\section{Introduction}
    \vspace{-1mm}

Visual understanding of scenes are broadly covered by object detection \cite{szegedy2013deep,papageorgiou2000trainable} and localization \cite{sullivan2001bayesian, conaire2007improved} for single or multiple objects. Evolved from detection of various objects, image segmentation \cite{cooper1998tractability, alvarez2012road} is another research topic which helps to understand the attributes of the scene. While these techniques supply some useful information of the image, but a scene also largely depends on the interactions or relations between objects. This idea led to scene graphs which describe the scene by incorporating the objects and their pairwise relations. This relation is represented by a directed edge pointing from the subject to the object. Evaluating a scene by detecting the objects and the relationships between them allows building a graph consisting of nodes representing the objects and the edges representing the relations. It consists of number of triplets represented in $<$subject-relationship-object$>$ form. They help in aiding deep understanding of the scene for various vision tasks such as visual reasoning \cite{shi2019explainable}, image captioning \cite{yang2019auto, li2019know}, image retrieval \cite{johnson2015image, schuster2015generating, schroeder2020structured} and visual question answering \cite{zhang2019empirical,ghosh2019generating,teney2017graph}. To improve these applications and their benefits, it is crucial to have a well performing model which generates scene graphs that corresponds to the actual visual scene. 

The directional nature of a triplet in scene graph defines the subject and object in a triplet. Scene graph generation (SGG) is considered a complex problem in computer vision because of the imbalanced nature of the datasets and intricate relationship information. As stated in MOTIFS \cite{zellers2018neural}, combinations of at least two triplets appear in many images. So, the presence of some objects highly increase the probability of the presence of other objects. Communication i.e. information flow between the detected objects has been shown to be beneficial for improving the performance of scene graph generation models \cite{zellers2018neural, chen2019counterfactual}.  

\begin{figure}
   \includegraphics[scale=0.33]{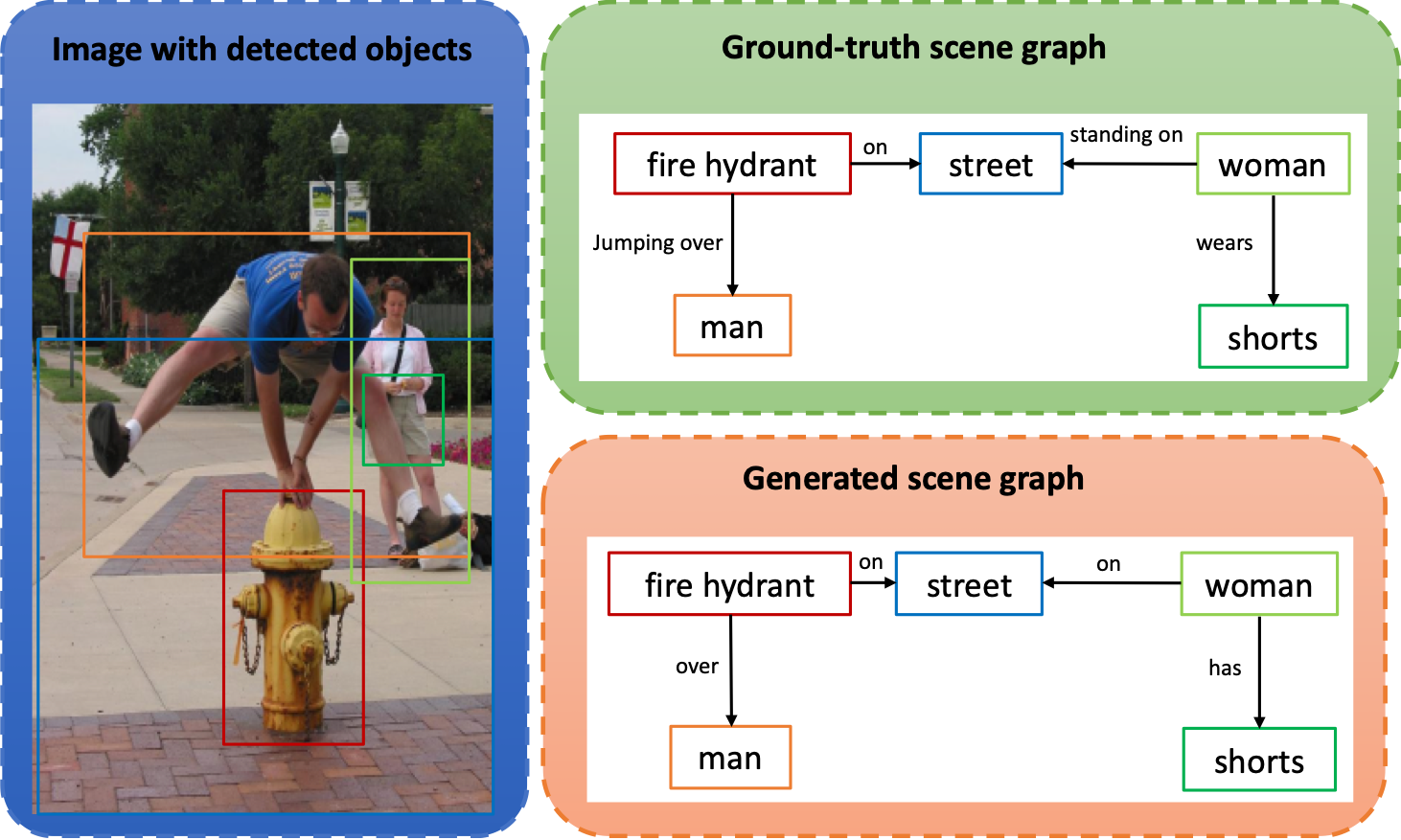}
    \caption{Two different scene graphs of the same given image on the left. Top: The ground-truth scene graph as annotated in the Visual Genome dataset. Below: A generated scene graph. As in many cases, the generated scene graph does capture the visual scene correctly, since the scene graph generation model only predicts frequently appearing relationships.}
    \vspace{-7mm}
    \label{sgg_1}
\end{figure}

The frequency distribution of the relationship within the Visual Genome dataset is long-tailed \cite{KrishnaZGJHKCKL16}. Due to this, scene graph detection models can already achieve good performance by only predicting the most frequent relationship for the respective subject-object pair. Figure \ref{sgg_1} illustrates the problem almost every SGG model faces. In many object pairings, the relationship is trivial and mostly possessive or geometric (e.g. \textit{on}, \textit{under}, or \textit{next to}). The detailed descriptive and semantic relationships such as \textit{jumping over} as shown in Figure \ref{sgg_1} (given in the ground-truth) will not often be predicted since it rarely occurs. For creating models that represent less frequent relationships more precisely, an approach to handle dataset bias must be found. Since the bias in the dataset can also be beneficial, e.g. the probability for the relationship \textit{reading} will be much higher than for \textit{eating} if the subject-object pair given is person and book \cite{tang2020unbiased}, traditional debiasing methods will most likely strongly harm the performance of the model. For this reason, the handling of the bias in the dataset is one of the most under-explored properties of the scene graph generation task. 

In this paper, we propose a novel model for SGG. The objects present in an image are highly dependent on the presence of other objects, for instance, if there is a bike in the scene, then there will be two tyres in the same scene with a very high probability. This model uses a bidirectional GRU (BiGRU) layer to send information from every object to every other. This allows benefiting from the fact that some objects will increase the possibility for specific other objects to be present. 

Subsequently to this layer that covers the object-object communication, a transformer encoder layer is used to predict the object classes. Objects and their preferred or observed relations are closely connected. To extract the information for the edge, a similar approach as in \cite{zellers2018neural} is followed to specify the edge context for every detected object. We use an additional transformer encoder layer for this task of extracting the edge context features.

Using the object representations, their respective edge context is then used for the relationship prediction. In this procedure, a log-softmax function is applied to the subject-object pairwise relationship distribution. Following this Frequency Softening (FS), a Bias Adaptation (BA) approach \cite{lin2020gpsnet} is used. The bias for every subject-object is controlled by the bias adaptation term which takes scene-specific inputs to vary the amount of added bias. 

The contribution summary of the proposed BGT-Net is given in four modules to improve scene graph generation performance: (1) Object-object communication is performed using the BiGRU's. (2) A transformer encoder with scaled-dot-product attention is used to predict object classes after they have received information about the other objects present in the scene. (3) For every object, a second transformer encoder is used to gather information for the edges. (4) To tackle the bias in the relationship distribution, FS and BA \cite{lin2020gpsnet} is adopted. 

The evaluation efficacy of the proposed BGT-Net is performed on three SGG datasets: Visual Genome (VG) \cite{KrishnaZGJHKCKL16}, OpenImages (OI) \cite{abs-1811-00982}, and Visual Relationship Detection (VRD) \cite{lu2016visual}. We perform extensive experiments and ablation study to demonstrate the effectiveness of BGT-Net. Experimental results illustrate that the proposed BGT-Net outperforms the state of the art SGG results on a common metric Recall@K and on different datasets to the best of our knowledge. 
\vspace{-2.5mm}

\begin{figure*}[tbp]
\setlength\abovecaptionskip{-1.8\baselineskip}
\setlength\belowcaptionskip{2pt}
\begin{center}
\vspace{-3mm}
\includegraphics[height=5cm, width=17.5cm]{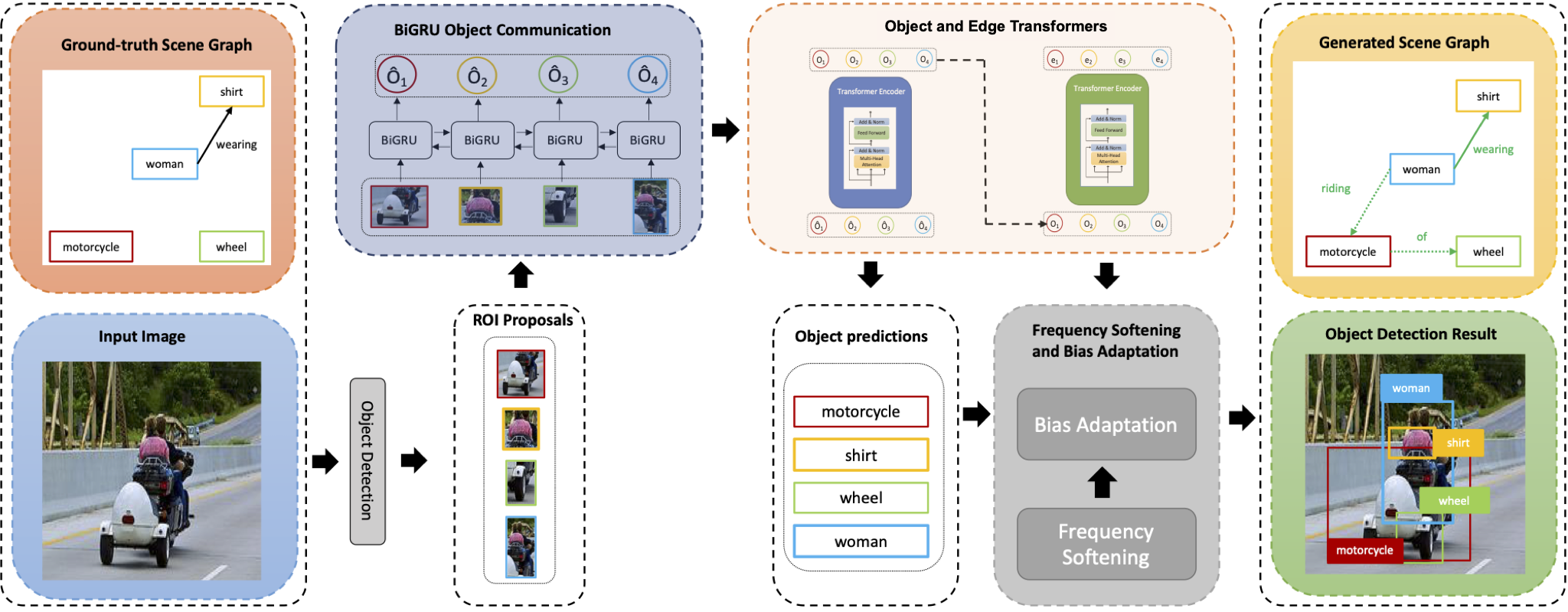}
\vspace{0mm}
\topcaption{The framework of BGT-Net uses Faster R-CNN (with VGG-16 or ResNext-101 as the backbone) to get the visual features and spatial locations of object proposals. It includes various sub-modules for the task of SGG: (1) New technique of using BiGRU for object-object communication, (2) Novel method of using a transformer encoder with scaled-dot-product attention for predicting  object classes after they have received information of the other objects present in the scene, (3) Additional transformer encoder is used to get the edge features, (4) FS and BA are used for dealing with the bias in the dataset. }
	\vspace{-1mm}
  \label{BGT-Model}
\end{center}
\end{figure*}

\vspace{-0.075mm}

\section{Related Work}
\vspace{-1.5mm}

From the ongoing research in scene graph generation, two different approaches for scene graph construction have developed. In the less common two-stage approach \cite{DBLP:journals/corr/abs-1812-01880, herzig2018mapping, chen2019counterfactual, DBLP:journals/corr/abs-1812-01880}, attributes of the scene graph are used in the second training step to refine the results produced by the first stage. Much more common are the one-stage approaches \cite{chen2019counterfactual,zellers2018neural,chen2019knowledge,DBLP:journals/corr/XuZCF17, yang2018graph, lin2020gpsnet,li2017scene,lu2016visual,li2018factorizable, qi2019attentive} which focus only on object detection and relationship classification, while almost neglecting intrinsic features. The proposed BGT-Net follows a one step approach and has the following advantages as compared to the literature work: (1) It uses object-object communication which improves the performance in SGG; (2) It deploys transformer encoder for object and edge context prediction which has shown to be highly beneficial in optimizing the parameters of SGG; (3) It is easy to train.

The MOTIFS \cite{zellers2018neural} stated in the early days of scene graph generation, that there are different combinations of triplets that appear in a lot of images. Therefore, a dependency between object appearances is present in the datasets. To leverage this information object communication has been examined in the CMAT model \cite{chen2019counterfactual} and improved the performance of the model. All of the published works show difficulties with the bias present in the Visual Genome dataset, which is widely used in the scene graph generation task. This bias arrives from the long-tailed relationship distribution. The GPS-Net \cite{lin2020gpsnet} tackled this problem with FS and BA which worked well compared to the previous works. The overall performance of the model could be improved as well as improvements in mean Recall@K were achieved, which gives reasoning about the positive effect of their approach in handling the dataset bias. So, motivated from GPS-Net, BGT-Net uses FS and BA. \break
While the GPS-Net changed the way the model was built, another recent work \cite{tang2020unbiased} developed a Scene Graph Diagnosis toolkit that can be used on a casually built scene graph. This tool kit is based on casual inference. Drawing the counterfactual causality to the proposed graph allows inferring with the bad bias. This approach did largely improve the mean Recall@K but decreased the other metrics significantly.  Similarly \cite{yan2020pcpl}, adaptively changed the weights of the loss by using the correlation between the relationship classes. This work improved mean Recall@K but had overall quite low Recall@K results.

\vspace{-1mm}

\section{Approach}
\label{chp:GRU-Transformer}

The illustration of the BGT-Model can be found in Figure \ref{BGT-Model}. The regions of interest and object proposals are obtained by employing a Faster R-CNN object detector \cite{ren2015faster}. Following the VG-split \cite{DBLP:journals/corr/XuZCF17}, there are 151 object categories (including 'background') and 50 relationship categories (including 'no relation'). For every proposal $i$, the visual feature $\boldsymbol{\hat{x_{i}}}$ is formed by concatenating the ROI (region of interest) feature $\boldsymbol{f_{i}} \in \mathbb{R}^{2048}$, the class confidence scores $\boldsymbol{s_{i}} \in \mathbb{R}^{151}$, and the spatial feature of the proposal bounding box $\boldsymbol{s_{i}} \in \mathbb{R}^{4}$. For the next step, the $\boldsymbol{x_{i}} \in \mathbb{R}^{512}$ gets transformed by the projection of $\boldsymbol{\hat{x_{i}}}$ into a 512-dimensional subspace. For the relationship classification, the union feature $\boldsymbol{u_{i,j}} \in \mathbb{R}^{2048}$ for every pair of objects $i$ and $j$ is extracted in the object detection stage. These feature vectors representing the scene are used in the following modules of the BGT-Model. In Section \ref{subsec:obj_com}, the object communication step  implemented with a BiGRU is explained in detail. Section \ref{chp:transformer}, introduces the object classification and edge information generation step using transformers. In Section \ref{chp:soft_adapt}, the FS of the long-tailed relationship distribution and the BA as a pre-processing for the relationship classification is described. 
 
\subsection{Object Communication}
\label{subsec:obj_com}

The object communication module takes the visual features $\boldsymbol{x_{i}}$ as input. The communication between the objects is implemented by a BiGRU. Due to the architecture of BiGRU, information from every object can flow to every other object. This information flow can be regulated by the BiGRU by learning which information shall be passed and which information shall be blocked. The output of the object communication step is therefore given by:
\begin{equation}
    \hat{O} = BiGRU([x_{i}]_{i=1,2,...,n}) 
\end{equation}
Where $\hat{O} = \{\hat{o}_{1}, \hat{o}_{2}, ... , \hat{o}_{n} \}$ are the object features after the communication step. The $\hat{o}_{i} \in \mathbb{R}^{1028}$ is obtained by concatenating the outputs $\overrightarrow{\hat{o}}_{i}$ (left to right in BiGRU) and $\overleftarrow{\hat{o}}_{i}$ (right to left in BiGRU), such that: 
	\vspace{-2mm}
\begin{equation}
   \hat{o}_{i} = [\overrightarrow{\hat{o}}_{i}, \overleftarrow{\hat{o}}_{i}] \in \mathbb{R}^{1028}
\end{equation}
	\vspace{-5mm}

\subsection{Object and Edge Transformers}
\label{chp:transformer}
The output $\hat{O}$ is projected to a 512-dimensional subspace to be fed into the Object transformer encoder. This transformer encoder follows the model architecture proposed by \cite{DBLP:journals/corr/VaswaniSPUJGKP17}, and takes the encoder block of the complete transformer model. This transformer encoder is built up by a Multi-Head attention layer, an Add \& Norm layer, a Feed Forward layer, and another Add \& Norm layer. 

The three inputs to the Multi-Head Attention layer are the values $V$, the keys $K$ and the queries $Q$. According to \cite{DBLP:journals/corr/VaswaniSPUJGKP17}, these three inputs are obtained from a single input. Using three different feed-forward fully-connected layers, yields the queries, keys, and values. For every attention head $i$ (here i=1,2,..,8) these values are calculated by:
	\vspace{-2mm}
\begin{equation}
    Q_{i} = \hat{O} * W^{Q}_{i}
\end{equation}
	\vspace{-5mm}
\begin{equation}
    K_{i} = \hat{O} * W^{K}_{i}
\end{equation}
	\vspace{-4mm}
\begin{equation}
    V_{i} = \hat{O} * W^{V}_{i}
\end{equation}
	\vspace{-1mm}
Where $W^{Q}_{i} \in \mathbb{R}^{512 \times d_{k}}$, $W^{K}_{i} \in \mathbb{R}^{512 \times d_{k}}$ and $W^{V}_{i} \in \mathbb{R}^{512 \times d_{v}}$ are learnable parameter matrices. Also, $d_{k} = d_{v} = 64$ are the same for this application. From these values, the Scaled Dot-Product attention is calculated, such that the output of each attention head is given by:
    \vspace{-3mm}
\begin{equation}
    Z_{i} = softmax(\frac{Q_{i} * K_{i}^{T}}{\sqrt{d_{k}}}) * V_{i}
\end{equation}
    \vspace{-1mm}
Concatenating $Z_{i}$'s give the output of the Multi-Head Scaled Dot-Product layer as $Z$. This is then fed through another fully-connected layer to bring it back to the dimension of the input matrix $\hat{O}$.
    \vspace{-3mm}
\begin{equation}
    Z = concatenate(Z_{1}, Z_{2}, ... , Z_{n_{heads}})
\end{equation}
    \vspace{-1mm}
The Add \& Norm layer adds the input of the previous Multi-Head Attention layer as a residual connection to the output of the attention layer. The normalization applied is a normalization layer following the approach of \cite{ba2016layer}. Here, it is suggested that the `covariate shift' problem can be reduced by changing the mean and the variance of the summed inputs in every layer. This is followed by a Feed Forward layer with two linear transformations and a ReLU activation followed by another Add \& Norm layer. This transformer encoder block gets repeated 6 times. The output $Z_{6}$ of the last repetition is then used to predict the object labels:
    \vspace{-3mm}
\begin{equation}
    O = softmax(W_{o} * Z_{6}')
\end{equation}
    \vspace{-1mm}
Where $W_{o} \in \mathbb{R}^{512 \times 151}$ predicts the object class distribution for each detected object. Repeating this procedure but using the output $Z_{6}$ of the Object transformer as input to the Edge transformer yields feature vectors having information about the edges for every object. Similarly to \cite{zellers2018neural}, this information then can be used in the relationship prediction step. This edge information is given formally by:
    \vspace{-3mm}
\begin{equation}
    E = TransformerEncoder(Z_{6})    
\end{equation}
    \vspace{-1mm}
Where $E = \{e_{1}, e_{2},..., e_{n}\}$ contains the edge information for every object. 

\subsection{Frequency Softening(FS), Bias Adaptation(BA)}
\label{chp:soft_adapt}

To handle the long-tailed relationship distribution present in the Visual Genome dataset, the procedure of softening this distribution and adapting the bias term for every subject-object pair form \cite{lin2020gpsnet} is adopted. The used features in this step is different than in the GPS-Net \cite{lin2020gpsnet} , but the principle stays the same. The softening of the relationship distribution is done by applying a log-softmax function to the original distribution. The softened frequency distribution is therefore given by Eq. \ref{eq:10}. The probability distribution $p_{i \rightarrow j}$ of every object pair $i$ and $j$ must be softened separately. Based on this definition, softening does not take any information of the respective visual scene into account.
 \vspace{-3mm}
\begin{equation} \label{eq:10}
     \tilde{p}_{i \rightarrow j} = log\ softmax(p_{i \rightarrow j})
     \end{equation}
    \vspace{-4mm}

 BA is used to get a case-specific adaptation of the above term. BA allows us to adjust the bias in the relationship prediction step. This adaptation term takes the appearance of the subject-object pair $i$, $j$ into account, by using their union feature $u_{i,j}$. The BA term $d$ can be calculated by:
    \vspace{-3mm}
\begin{equation}
    d = W_{p} * u_{i,j}
\end{equation}
    \vspace{-1mm}
Where $W_{p} \in \mathbb{R}^{2048}$ is the transformation matrix and $u_{i,j}$ is the union feature of the subject-object pair. 
In the relationship prediction step, this bias term can be used as follows:
    \vspace{-3mm}
\begin{equation}
    p_{i,j} = softmax( W_{r}(o'_{i} * o'_{j} * u_{i,j}) + d \odot \tilde{p}_{i \rightarrow j}  )  
\end{equation}
    \vspace{-0.5mm}
In this equation, the bias $d \odot \tilde{p}_{i \rightarrow j}$ can be adjusted by changing $d$ accordingly. Here, $o'_{i} = [o_{i}, e_{i}]$ (same for $o'_{j}$) with '[-,-]' denotes the concatenation function representing the object features obtained in earlier steps. $W_{r}$ is the classifier that projects the features to the relationship class dimension. $ \odot $ denotes the Hadamard Product and $ * $ the fusion function. The fusion function for $ (x * y)$ is given by:
    \vspace{-3mm}
\begin{equation}
    (x*y) = (W_{x}x + W_{y}y) - (W_{x}x - W_{y}y) \odot  (W_{x}x - W_{y}y)
\end{equation}
    \vspace{-1mm}
With the parameter matrices $W_{x}$ and $W_{y}$, the fusion function is proposed to learn to count objects in images. \cite{DBLP:journals/corr/abs-1802-05766}.

The predicted relationship between objects $i$ and $j$ is given by:
\begin{equation}
    r_{i,j} = argmax(p_{i,j}(r)) 
\end{equation}
Where $r$ lies in the set of relationship classes including background BG.

 \vspace{-1mm}
\section{Experiments}
 \vspace{-1mm}
We performed experiments using three different datasets, i.e., Visual Genome (VG) \cite{KrishnaZGJHKCKL16}, Visual Relationship Detection (VRD) \cite{lu2016visual}, and OpenImages (OI) \cite{abs-1811-00982}. 
 \vspace{-1mm}
\subsection{Visual Genome}
 \vspace{-1mm}
Visual Genome dataset \cite{KrishnaZGJHKCKL16} is the most frequently used dataset for the SGG task. We use the same data statistics and evaluation metrics as widely used by the state of the art in this field, i.e., 150 object categories and 50 relationship categories are used. 70\% of the dataset is used for training and 30\% for testing. An additional 5000 images are taken from the training set and are used for validation. As used by the state of the art, we also employed Faster R-CNN \cite{ren2015faster} with VGG-16 or ResNext-101 as a backbone to get the characteristics of object proposals. To keep the fairness in the comparison with state-of-the-art, we chose same experimental factors as chosen by \cite{lin2020gpsnet}.

\textbf{Performance Diagnosis:} The scene graph generation model is evaluated in three different sub-tasks: (1) Predicate classification, (2) Scene graph classification, and (3) Scene graph generation. These are the three protocols for which the model's performance is evaluated separately. 
Predicate Detection is used to specify the relation of given objects. This protocol evaluates the set of possible relations between a pair of given objects. The prediction of relationships without the effect of object detection is examined.
In Phrase Detection, the input is an image and the outputs are triplets of subject-predicate-object. Additionally, one bounding box must have an overlap of at least 0.5 with the corresponding ground truth.
For Relation Detection, the same input and output as in Phrase Detection is used. In this case, not only one but two bounding boxes of the pair of objects must have at least 0.5 overlap with the ground truth. 

\textbf{Metrics.} The evaluated metrics for the diagnosis of the model performance is Recall at K (R@K), no graph constraint Recall at K (nGR@K), zero-shot Recall at K (zsR@K), and mean Recall (mR@K), where $K$=20, 50, and 100, respectively.

\textbf{Object Detector:} A pretrained Faster R-CNN object detector with a VGG-16 net \cite{han2015learning} or ResNeXt-101-FPN \cite{xie2017aggregated} as a backbone is used which is taken from \cite{lin2020gpsnet}. This detector was trained on the VG dataset, with batch size 8 and initial learning rate $8*10^{-3}$ which is decayed at the $30k^{th}$ and $40k^{th}$ iteration by the factor of 10. After training this detector on 4 2080Ti GPU, 28.14 mAP (with 0.5 IoU) was achieved. 

\textbf{Scene Graph Generation:} The scene graph generation is trained with an SGD optimizer \cite{bottou2012stochastic}. The learning rate is set at $10^{-3}$ for all three protocols. This learning rate was decayed by a factor of 10 twice after hitting a validation performance plateau. Per-Class non-maximal suppression (NMS) was applied with 0.5 IoU. 160 RoIs for each image were sampled. In contrast to previous works, we also considered non-overlapping regions for relationship prediction. To generalize the scene graph generation task, we used similar settings as used in literature \cite{lin2020gpsnet}. For model training, an RTX Titan was used. The batch size was set to 12 and the learning rate started at $10^{-3}$ and was reduced two times by a factor of 10 after hitting a validation performance plateau. The number of solver iterations was set to 18000. 
 \vspace{-3mm}
\subsection{OpenImages}
 \vspace{-1mm}
The training and validation sets of the OpenImages dataset contain 53,953 and 3,234 images. For comparison, we use the same Faster R-CNN detector with a pre-trained ResNeXt- 101-FPN backbone as used by \cite{lin2020gpsnet,zhang2019graphical}. Also, the same data processing and evaluation metrics are used as in these previous works \cite{lin2020gpsnet,zhang2019graphical}. The evaluation metrics are Recall$@$50, weighted mean average precision (AP) of relationships wmAP\textsubscript{rel}, and weighted mean AP of phrase wmAP\textsubscript{phr}. The final score is given by $0.2*R@50 + 0.4*wmAP_{rel} + 0.4*wmAP_{phr}$, which was adopted from the OpenImages challenge formula \cite{DBLP:journals/corr/abs-1903-02728}, where the mAP was replaced by its weighted counterpart. The replacement of the mAP with the wmAP was done \cite{DBLP:journals/corr/abs-1903-02728} due to the extreme predicate class imbalance. The wmAP is achieved by scaling each predicate category by their relative ratios in the val set from the mAP. Important to note is that the wmAP\textsubscript{rel} evaluates the AP of the predicted triplet where both the subject and object boxes have an IoU of at least 0.5 with ground truth. The wmAP\textsubscript{phr} is quite similar but is utilized for the union box of the subject and the object. 
 \vspace{-2mm}
\subsection{Visual Relation Detection}
 \vspace{-1mm}
The Visual Relation Detection (VRD) dataset was introduced by \cite{lu2016visual}. We adopt the same detectors as \cite{DBLP:journals/corr/abs-1903-02728}. Specifically, we use a pre-trained Faster R-CNN detector with VGG-16 backbone trained on the COCO dataset \cite{lin2014microsoft}. The evaluation process is given by \cite{lu2016visual} and the metrics used are R$@$50 and R$@$100.

\subsection{Implementation Details}
We follow the same implementation parameters as used by \cite{lin2020gpsnet}. To ensure a fair comparison with state of the art, we used VGG-16 and ResNext-101 as backbone. We use the 10$^{-3}$ as the learning rate and 6 as the batch-size which is the same as used by \cite{lin2020gpsnet}. We use SGD with momentum as the optimizer for the training process. We use the relationship between overlapped bounding boxes and subject-object pairs for the SGDet process. NMS with an IoU of 0.3 is used and the topmost 64 object proposals are chosen. The ratio of 3:1 is maintained during training between the subject-object pairs with and without any relationship. 

\subsection{Comparisons with State-of-the-Art Methods}

\begin{table*}[!htb]
	\begin{minipage}{\textwidth}
		\centering
		\footnotesize
		\resizebox{\columnwidth}{!}{%
		\begin{tabular}{llccccccccccccc}
		\hline
		  &  &  & SGdet &  &  &  & SGCls &  &  &  & PredCls & & \\
		 \hline
		 \textbf{Model} & \vline & $R@20$  & $R@50$ & $R@100$ & \vline & $R@20$  & $R@50$ & $R@100$ & \vline & $R@20$  & $R@50$ & $R@100$& \vline & Mean \\
    	 \hline
			MOTIFS $\star$ \cite{zellers2018neural} & \vline & 21.4  & 27.2 & 30.3 & \vline & 32.9  & 35.8 & 36.5 & \vline & 58.5  & 65.2 & 67.1 & \vline & 43.7\\
			FREQ $\star$ \cite{zellers2018neural} & \vline & 20.1  & 26.2 & 30.1 & \vline & 29.3  & 32.3 & 32.9 & \vline & 53.6  & 60.6 & 62.2 & \vline & 40.7\\
			VCTREE-SL $\star$ \cite{DBLP:journals/corr/abs-1812-01880} & \vline & 21.7  & 27.7 & 31.1 & \vline & 35.0  & 37.9 & 38.6 & \vline & 59.8  & 66.2 & 67.9 & \vline & 44.9\\
			VCTREE-HL $\star$ \cite{DBLP:journals/corr/abs-1812-01880} & \vline & 22.0  & 27.9 & 31.3 & \vline & 35.2  & 38.1 & 38.8 & \vline & 60.1  & 66.4 & 68.1 & \vline & 45.1\\
			GB Net $\star$ \cite{Zareian_2020_ECCV} & \vline & -  & 26.3 & 29.9 & \vline & -  & 37.3 & 38 & \vline & -  & 66.6 & 68.2 & \vline & 44.4\\
			NODIS $\star$ \cite{yuren2020nodis} & \vline & 21.5  & 27.4 & 30.7 & \vline & 36  & 39.8 & 40.7 & \vline & 58.9  & 66 & 67.9 & \vline & 45.4\\
			GPS-Net $\star$ \cite{lin2020gpsnet} & \vline & 22.6  & 28.4 & 31.7 & \vline & 36.1  & 39.2 & 40.1 & \vline & 60.7  & 66.9 & 68.8 & \vline & 45.9\\
			CMAT $\star$ \cite{chen2019counterfactual} & \vline & 22.1  & 27.9 & 31.2 & \vline & 35.9  & 39 & 39.8 & \vline & 60.2  & 66.4 & 68.1 & \vline & 45.4\\
			KERN $\star$ \cite{chen2019knowledge} & \vline & -  & 27.1 & 29.8 & \vline & -  & 36.7 & 37.4 & \vline & -  & 65.8 & 67.6 & \vline & 44.1 \\
			Graph R-CNN $\star$  \cite{yang2018graph} & \vline & -  & 11.4 & 13.7 & \vline & -  & 29.6 & 31.6 & \vline & -  & 54.2 & 59.1 & \vline & 33.3\\
			IMP $\star$ \cite{DBLP:journals/corr/XuZCF17} & \vline & -  & 3.44 & 4 .24 & \vline & -  & 21.72 & 24.38 & \vline & -  & 44.75 & 53.08 & \vline & 25.3\\
			\hline
			\textbf{BGT-Net (no BiGRU)} $\star$ & \vline & \textbf{23.61}	& \textbf{ 30.4}	& \textbf{  34.81} 	& \vline & \textbf{ 33.81}	& \textbf{	37.22}	& \textbf{	38.12}	& \vline & \textbf{ 57.98} & \textbf{ 64.75} & \textbf{ 66.63} & \vline & \textbf{46.9}\\
			\hline
			\textbf{BGT-Net} $\star$ & \vline & \textbf{23.1}	& \textbf{28.6}	& \textbf{32.2} 	& \vline & \textbf{38.0}	& \textbf{40.9}	& \textbf{43.2}	& \vline & \textbf{60.9}	& \textbf{67.3}	& \textbf{68.9} & \vline & \textbf{46.9}\\
			\textbf{BGT-Net} $\diamond$ & \vline & \textbf{25.5}	& \textbf{32.8}	& \textbf{37.3} 	& \vline & \textbf{41.7}	& \textbf{45.9}	& \textbf{47.1}	& \vline & \textbf{60.9}	& \textbf{67.1}	& \textbf{68.9} & \vline & \textbf{49.9}\\

		\hline
  	\end{tabular}}
  	\vspace{0mm}
  	\normalsize
	{\topcaption[Recall@K Model comparison with state-of-the-arts on the VG dataset.]{Recall@K Model comparison with state-of-the-arts on the VG dataset. We compare R@20, R@50, and R@100. For some literature work, the R@20 is not given. We used '$-$'  which denotes that the result is unavailable. So, the mean is calculated using values of R@50 and R@100 to have fair comparison with state of the art. Models using the same VGG backbone are denoted with '$\star$' and the BGT-Net with ResNext-101 background is marked with '$\diamond$'.}\label{tab:comp_modelsVG}}
		\end{minipage} \hfill
\vspace{-8mm}
\end{table*} 

\begin{table}[h]
		\centering
		\footnotesize
		\resizebox{\columnwidth}{!}{%
		\begin{tabular}{llccccccc}
		\hline
		 &  & SGdet &  &   SGCls  &  & PredCls & &  \\
		 \hline
		 \textbf{Model} & \vline &  $mR@100$ & \vline & $mR@100$ & \vline & $mR@100$ & \vline & Mean \\
    	 \hline
			GB Net $\star$ \cite{Zareian_2020_ECCV} & \vline & 8.5 & \vline & 13.4 & \vline & \textbf{24} & \vline & 15.3 \\
		    
		    IMP $\star$ \cite{DBLP:journals/corr/XuZCF17} & \vline & 4.8 & \vline & 6.0 & \vline & 10.5 & \vline & 7.1 \\
		    	
			GPS-Net $\star$ \cite{lin2020gpsnet} & \vline & \textbf{9.8} & \vline & 12.6 & \vline  & 22.8 & \vline & 15.1 \\
			
			VCTREE-HL $\star$ \cite{DBLP:journals/corr/abs-1812-01880} & \vline  & 8.0 & \vline  & 10.8 & \vline & 19.4 & \vline & 12.8 \\
			
			MOTIFS $\star$ \cite{zellers2018neural} & \vline  & 6.6 & \vline  & 8.2 & \vline & 15.3 & \vline & 10.0\\

			KERN $\star$ \cite{chen2019knowledge} & \vline  & 7.3 & \vline  & 10 & \vline & 19.2 & \vline & 12.2\\
			
			\hline
			\textbf{BGT-Net} $\star$ & \vline & 9.6 	& \vline & \textbf{13.7}	& \vline 	& 23.2 & \vline & \textbf{15.5}\\
		
		\hline
  	\end{tabular}}
  	\normalsize
  		\vspace{-2mm}
	\caption[Comparison on mR@100]{Comparison on mR@100 between various methods across all 50 relationship categories.} \label{tab:meanRecall}
		\vspace{-7mm}
\end{table}

\begin{table*}[!htb]
	\begin{minipage}{\textwidth}
		\centering
		\footnotesize
		\resizebox{\columnwidth}{!}{%
		\begin{tabular}{llccccccccccc}
		\hline
		  &  &  & SGdet &  &  &  & SGCls &  &  &  & PredCls &  \\
		 \hline
		 \textbf{Model} & \vline & $nGR@20$  & $nGR@50$ & $nGR@100$ & \vline & $nGR@20$  & $nGR@50$ & $nGR@100$ & \vline & $nGR@20$  & $nGR@50$ & $nGR@100$ \\
    	 \hline
		
			GB Net $\star$ \cite{Zareian_2020_ECCV} & \vline  & - & 29.3	& 35 & \vline & - & 46.9	& 50.3 & \vline	& - &	\textbf{83.5} & \textbf{90.3}\\
			
			CMAT $\star$ \cite{chen2019counterfactual} &\vline & 23.7	& 31.6	& 36.8 & \vline & \textbf{41}	& \textbf{48.6}	& \textbf{52} & \vline & \textbf{68.9}	& 83.2	& 90.1\\
			KERN $\star$ \cite{chen2019knowledge} &\vline & - & 30.9	& 35.8  & \vline & - & 45.9	& 49  &\vline & - & 65.8	& 67.6\\
			\hline
			\textbf{BGT-Net (no BiGRU)} $\diamond$ & \vline  & 24.23	& 32.88	& 39.06 & \vline & 38.55	& 46.28	& 50.04 & \vline & 65.92 & 80.51	& 87.82 \\
				
			\textbf{BGT-Net} $\diamond$ & \vline & \textbf{27.24}	& \textbf{36.91}	& \textbf{43.72} 	& \vline & \textbf{47.83}	& \textbf{57.67}	& \textbf{62.29}	& \vline & \textbf{69.1}	& \textbf{83.71}	& \textbf{90.55} \\
		\hline
		 \hline
				 \textbf{Model} & \vline & $mR@20$  & $mR@50$ & $mR@100$ & \vline & $mR@20$  & $mR@50$ & $mR@100$ & \vline & $mR@20$  & $mR@50$ & $mR@100$ \\
    	 \hline
			GB Net $\star$ \cite{Zareian_2020_ECCV} & \vline & -  & 7.1 & 8.5 & \vline & -  & 12.7 & 13.4 & \vline & -  & \textbf{22.1} & \textbf{24} \\
			
			GPS-Net$\star$\cite{lin2020gpsnet} & \vline & -  & - & \textbf{9.8} & \vline & -  & - & 12.6 & \vline & -  & - & 22.8 \\
			
			KERN $\star$\cite{chen2019knowledge} & \vline & -  & 6.4 & 7.3 & \vline & -  & 9.4 & 10 & \vline & -  & 17.7 & 19.2 \\
			
			\hline
			\textbf{BGT-Net (no BiGRU)} $\diamond$ & \vline  & 4.62	& 6.55	& 7.85 & \vline & 7.31	& 9.14	& 9.71 & \vline & 12.14	& 15.59	& 17.05 \\
			\textbf{BGT-Net} $\diamond$ & \vline & \textbf{5.69}	& \textbf{7.81}	& 9.25 	& \vline & \textbf{10.41}	& \textbf{12.77}	& \textbf{13.61}	& \vline & \textbf{16.8}	& \textbf{20.56}	& \textbf{22.98} \\
		\hline
			 \hline
		 \textbf{Model} & \vline & $zsR@20$  & $zsR@50$ & $zsR@100$ & \vline & $zsR@20$  & $zsR@50$ & $zsR@100$ & \vline & $zsR@20$  & $zsR@50$ & $zsR@100$ \\
    	 \hline
			Motifs $\star$\cite{tang2020sggcode} & \vline & 0  & 0.05 & 0.11 & \vline & 0.32  & 0.91 & 1.39 & \vline & 1.35  & 3.63 & 5.36 \\
			
			IMP $\star$\cite{DBLP:journals/corr/XuZCF17} & \vline & 0.18  & 0.38 & 0.77 & \vline &  2.01 & 3.03 & 3.92 & \vline & 12.17  & 17.66 & 20.25 \\
			\hline
			\textbf{BGT-Net} $\diamond$& \vline & \textbf{1.22}	& \textbf{2.38}	& \textbf{3.42} 	& \vline & \textbf{4.8}	& \textbf{7.37}	& \textbf{8.78}	& \vline & \textbf{12.23}	& \textbf{18.31}	& \textbf{21.51} \\
		
		\hline
		
  	\end{tabular}}
  	\vspace{1mm}
  	\normalsize
	{\topcaption[No Graph Constraint Recall Results of transformer Model]{nGR@K, mR@K and  zsR@K comparison with state-of-the-art on the VG dataset. The version of the BGT-Net, i.e., the BGT-Net (no BiGRU) is  also compared.}\label{tab:comp_models_ngc1}}
		\end{minipage} \hfill
\vspace{-9mm}
\end{table*}

\begin{table}[h]
		\centering
		\footnotesize
		\resizebox{\columnwidth}{!}{%
		\begin{tabular}{llcccccccccccccc}
		\hline
		  & \vline & Predicate Detection  &\vline &  \multicolumn{2}{c}{Relation Detection}  &  &\multicolumn{2}{c}{Phrase Detection} &  & \\
		 \hline
		 \textbf{Model} & \vline & $R@50$  & \vline & $R@50$ & $R@100$ & \vline & $R@50$ & $R@100$ & \vline & Mean \\
    	 \hline
    	 VTransE \cite{DBLP:journals/corr/ZhangKCC17} & \vline & 44.8 & \vline & 19.4 & 22.4 & \vline & 14.1 & 15.2 & \vline & 23.2\\
    	 ViP-CNN \cite{article} & \vline & - & \vline & 17.3 & 20.0 & \vline & 22.8 & 27.9 & \vline & 22\\
         VRL \cite{DBLP:journals/corr/LiangLX17} & \vline & - & \vline & 18.2 & 20.8 & \vline & 21.4 & 22.6 & \vline & 20.8\\
         KL distilation \cite{DBLP:journals/corr/YuLMD17} & \vline & 55.2 & \vline & 19.2 & 21.3 & \vline & 23.1 & 24.0 & \vline & 28.6\\
         MF-URLN \cite{DBLP:journals/corr/abs-1905-01595} & \vline & 58.2 & \vline & 23.9 & 26.8 & \vline & 31.5 & 36.1 & \vline & 35.3\\
         Zoom-Net $\diamond$ \cite{DBLP:journals/corr/abs-1807-04979} & \vline & 50.7 & \vline & 18.9 & 21.4 & \vline & 24.8 & 28.1 & \vline & 28.8\\ 
         CAI + SCA-M $\diamond$ \cite{DBLP:journals/corr/abs-1807-04979} & \vline & 56.0 & \vline & 19.5 & 22.4 & \vline & 25.2 & 28.9 & \vline & 30.4\\ 
         RelDN $\star$ \cite{DBLP:journals/corr/abs-1903-02728} & \vline & - & \vline & 25.3 & 28.6 & \vline & 31.3 & 36.4 & \vline & 30.4\\ 
         GPS-Net $\star$ \cite{lin2020gpsnet} & \vline & 63.4 & \vline & 27.8 & 31.7 & \vline & 33.8 & 39.2 & \vline & 39.2\\ 
         \hline
         BGT-Net $\star$ & \vline & \textbf{64.1} & \vline & \textbf{28.5} & \textbf{31.9} & \vline & \textbf{34.4} & \textbf{39.4} & \vline & \textbf{39.6}\\ 
    	 
		\hline
  	\end{tabular}}
  	\normalsize
  	\vspace{-1mm}
	\caption[Comparison on VRD]{Comparison on the VRD dataset \cite{lu2016visual}. '$\star$' and '$\diamond$' denote the models using the same object detector. The object detector with a VGG-16 backbone trained on COCO is used as in RelDN and GPS-Net to have fair comparison.}
		\vspace{-7mm}
	\label{tab:comp_vrd}
\end{table}

\begin{table*}[h]
	\begin{minipage}{\textwidth}
		\centering
		\footnotesize
		\resizebox{\columnwidth}{!}{%
		\begin{tabular}{llcccccccccccccc}
		\hline
		  & \vline &  & &  &  & \vline  & \multicolumn{9}{c}{AP\textsubscript{rel} per class} \\
		 \hline
		 \textbf{Model} & \vline & $R@50$  & $wmAP_{rel}$ & $wmAP_{phr}$ &  $score_{wtd}$  & \vline & at & on & holds & plays  & interacts with & wears & hits & inside of & under\\
    	 \hline
		RelDN, $L_{0}$ \cite{DBLP:journals/corr/abs-1903-02728}& \vline & 74.67  & 34.63 & 37.89 & 43.94  & \vline & 32.40 & 36.51 & 41.84 & 36.04  & 40.43 & 5.70 & 55.40 & 44.17  & 25.00\\	
		RelDN \cite{DBLP:journals/corr/abs-1903-02728}& \vline & 74.94  & 35.54 & 38.52 & 44.61  & \vline & 32.90 & 37.00 & 43.09 & 41.04  & 44.16 & 7.83 & 51.04 & 44.72  & 50.00\\	
		GPS-Net \cite{lin2020gpsnet}& \vline & 77.29  & 38.78 & 40.15 & 47.03  & \vline & 35.10 & 38.90 & \textbf{51.47} & 45.66  & \textbf{44.58} & \textbf{32.35} & 71.71 & 47.21  & 57.28\\
		\hline
		BGT-Net & \vline & \textbf{77.98}  & \textbf{39.56} & \textbf{40.75} & \textbf{47.67}  & \vline & \textbf{36.23} & \textbf{39.05} & 50.96 & \textbf{46.78}  & 44.56 & 31.45 & \textbf{72.17} & \textbf{48.03}  & \textbf{57.64}\\
		\hline
  	\end{tabular}}
  	\normalsize
	\end{minipage} \hfill
	\vspace{-2mm}

	\caption[Comparison on OpenImages]{Comparison on OpenImages dataset \cite{DBLP:journals/corr/abs-1811-00982}. The BGT-Net uses a ResNext-101 backbone. Additionally, the same data processing and evaluation metrics as \cite{lin2020gpsnet, DBLP:journals/corr/abs-1903-02728} are followed to a ensure fair comparison.}
	\label{tab:comp_openimages}
	\vspace{-4mm}

\end{table*}

\textbf{Visual Genome:} 
BGT-Net outperforms all the compared previous literature work as shown in Table \ref{tab:comp_modelsVG} on R@K for all values of K . BGT-Net performs better than a recent model named GPS-net \cite{lin2020gpsnet} by 6$\%$  and by 12$\%$ on average at R@50 and R@100 over the three protocols when VGG-19 or ResNext-101 is used as the base for the Faster-RCNN, respectively. It also outperforms when an individual evaluation protocol is compared. The improvement is by 16.6$\%$ for SGDet, 17.3$\%$ for SGCls, and for 0.2$\%$ for PredCls. When compared to the classic MOTIFS \cite{zellers2018neural}, it showed an improvement of 7.3$\%$  and 14$\%$ on average at R@50 and R@100 over the three protocols when VGG-19 or ResNext-101 is used as a backbone for the Faster-RCNN, respectively. BGT-Net outperforms FREQ \cite{zellers2018neural}, VCTREE-SL \cite{DBLP:journals/corr/abs-1812-01880}, VCTREE-HL \cite{DBLP:journals/corr/abs-1812-01880}, GB-NET \cite{Zareian_2020_ECCV} , NODIS \cite{yuren2020nodis}, CMAT \cite{chen2019counterfactual}, KERN \cite{chen2019knowledge}, Graph R-CNN \cite{yang2018graph}, IMP \cite{DBLP:journals/corr/XuZCF17} by 14.9$\%$, 4.3$\%$, 3.9$\%$, 5.6$\%$, 3.2$\%$, 3.2$\%$, 6.3$\%$, 40.9$\%$, 85.4$\%$, respectively, on average at R@50 and R@100 over the three protocols when VGG-19 is used as a backbone for the Faster-RCNN and by 22.2$\%$, 11$\%$, 10.5$\%$, 12.3$\%$, 9.8$\%$, 9.8$\%$, 13.1$\%$, 49.9$\%$, 97.3$\%$, respectively, on average at R@50 and R@100 over the three protocols when ResNext-101 is used as a backbone for the Faster-RCNN.

\begin{figure}[ht]
\setlength\abovecaptionskip{0.4\baselineskip}
\setlength\belowcaptionskip{-18pt}
\begin{center}
\includegraphics[scale=0.3]{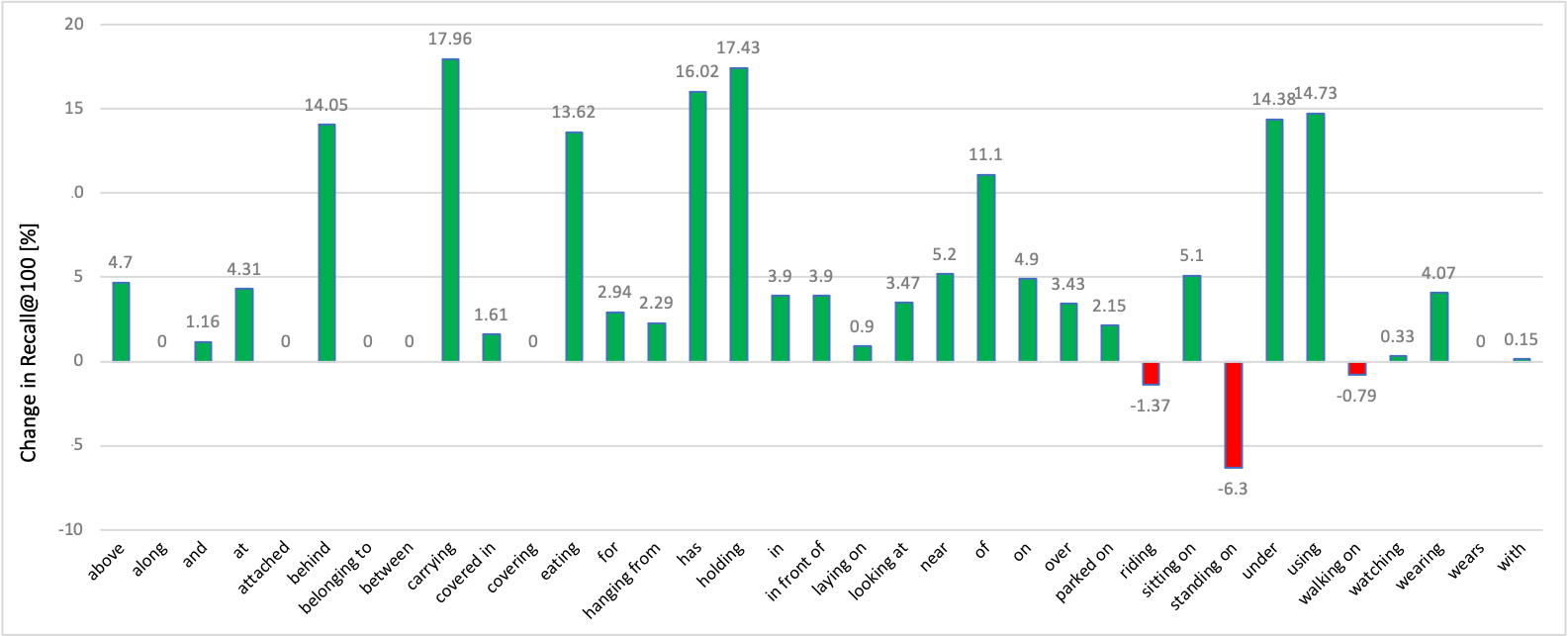}\\
\includegraphics[scale=0.3]{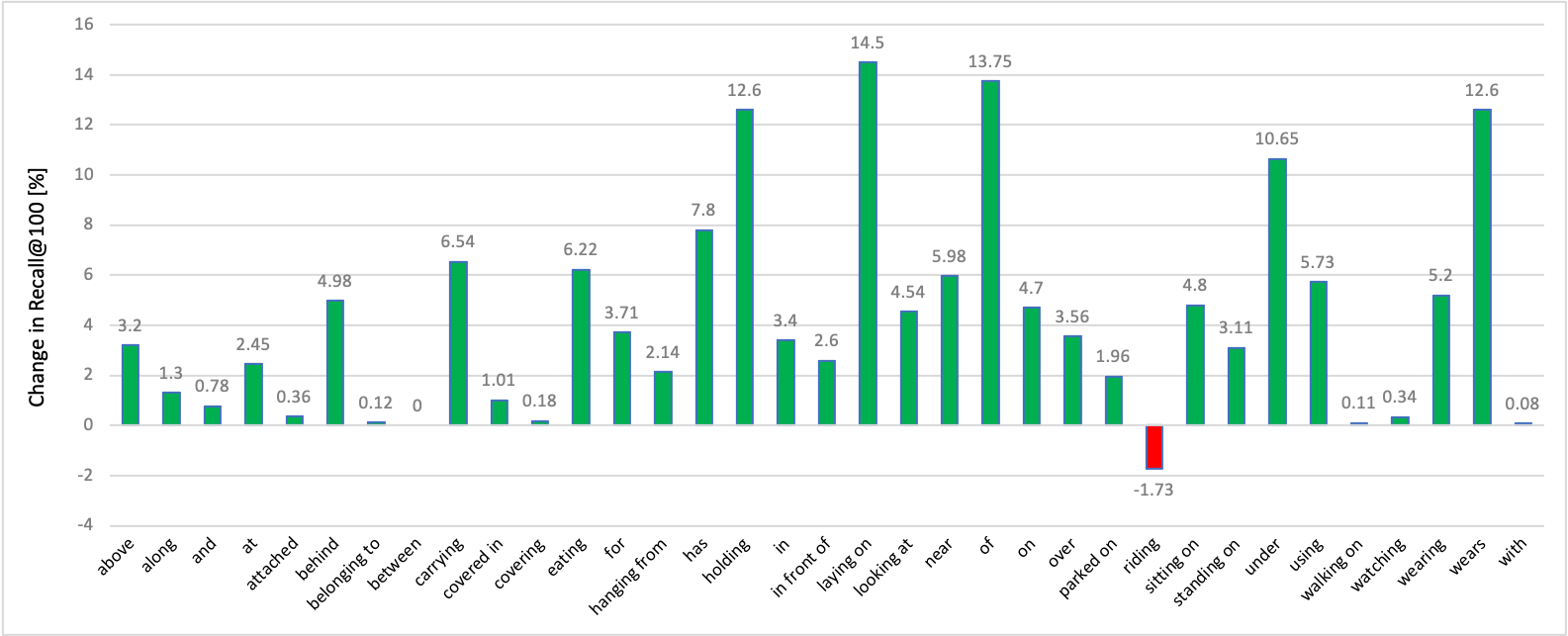}
\par\nointerlineskip\vspace{-1mm}
\caption{The increase in R@100 in PredCls of BGT-Net compared with the MOTIFS \cite{zellers2018neural} and VCTREE \cite{DBLP:journals/corr/abs-1812-01880}. The Top-35 relationship categories are selected according to their alphabetic occurrence.}
\vspace{-2.5mm}
\label{meanrecall12}
\end{center}
\end{figure}

We evaluate Mean Recall of BGT-Net to understand its performance on the class imbalance problem. So, we study its performance by conducting experiments to calculate its Mean Recall \cite{chen2019knowledge,DBLP:journals/corr/abs-1812-01880, lin2020gpsnet}. We see in Figure \ref{meanrecall12} and Table \ref{tab:meanRecall} that BGT-Net performs well considering the Mean Recall evaluation metric. The mean of the Mean Recall over all the three evaluation metrics (SGDet, SGCls, PredCls) is 15.5 for BGT-Net and hence outperforms GB-Net and GPS-Net which are the best state-of-the-art on class imbalance handling having good performance on both mean R@K and R@K results. This gives a positive indication that BGT-Net can tackle the problem of class imbalance while simultaneously giving high R@K as compared to the other existing solutions to best of our knowledge.

\textbf{Why BiGRU?} We investigated BGT-Net without BiGRU but BGT-Net (no BiGRU) has lower SGCls and PredCls results (see Table \ref{tab:comp_modelsVG}). The deciding factor for using BiGRU in BGT-Net is mR@K and nGR@K. Both these metrics significantly improve for all three: SGDet, SGCls, PredCls when BGT-Net is with BiGRU as shown in Table \ref{tab:comp_models_ngc1}.\\
Also shown in Table \ref{tab:comp_models_ngc1}, BGT-Net has high improvement on zsR@K as compared to \cite{DBLP:journals/corr/XuZCF17} and \cite{zellers2018neural} which shows that it is able to better detect those subject-predicate-object combinations which are not present in the training set. 

\vspace{-1mm}
\textbf{OpenImages:}
The results in Table \ref{tab:comp_openimages} show that BGT-Net performs very well on the OpenImages dataset. The overall score is 0.64 points higher than the GPS-Net model performance. This increase in performance is achieved by an overall increase in performance in all three evaluated metrics R$@$K, wmAP\textsubscript{rel}, and wmAP\textsubscript{phr}. An evaluation of the per-class AP is also shown. In this evaluation, some classes were chosen as in GPS-Net \cite{lin2020gpsnet} and RelDN \cite{DBLP:journals/corr/abs-1903-02728} to show class-specific performance. The performances of GPS-Net and the BGT-Net are quite close. For ''holds'', ''interacts with'' and ''wears'', the GPS-Net shows a higher AP while for the others the BGT-Net shows the highest AP.
The overall performance of the BGT-Net outperforms the state-of-the-art performance of the GPS-Net model. 

\textbf{Visual Relationship Detection:} 
The evaluation results on VRD Dataset are illustrated in Table \ref{tab:comp_vrd}. The BGT-Net uses the same Detector as RelDN and GPS-Net. The BGT-Net outperforms the state-of-the-art models in all three evaluation metrics. So, BGT-Net shows the best to the date performance on the VRD dataset as well to the best of our knowledge, and which is better than GPS-Net and RelDN (previous best performing networks).

\vspace{-2mm}
\subsection{Ablation Studies}
\vspace{-1mm}
To evaluate and analyze our proposed BGT-Net, we conducted a number of ablations as shown in Table \ref{tab:ablation1}, Table \ref{tab:ablation2}, and Table \ref{tab:ablation3}.

\textbf{Network Performance with a different combination of Modules.}
In this study, we evaluate the effectiveness of the network in the presence of the three modules, i.e., Transformer, BiGRU, and FS, individually and collectively.
The performance of the network increases with the presence of all three modules. Firstly, we evaluated for an individual module and then permutated the modules with each other to make a combination to evaluate the performance of different resulting configurations. As illustrated in Table \ref{tab:ablation1}, for all the three evaluation protocols, i.e., SGDet, SGCls, and PredCls, our proposed network with all the three modules outperforms the other network configurations with individual modules. When modules are used together, the network performance improves which shows that each individual module plays a significant role in predicting objects and their relationships.

The FS and BA were adapted from [20]. In \cite{lin2020gpsnet}, they performed ablation study with and without FS and BA. It was shown that these modules improve the PredCls for all three R@20, R@50, and R@100. We included in our ablation the effect of using FS which showed that zsR@20, zs@50, and zsR@100 improved drastically for all three: SGDet, SGCls, PredCl, when using FS as shown in Table \ref{tab:ablation1}.

\textbf{Performance with different number of Transformer Heads.}
We performed this ablation to validate the optimized number of transformer heads in the network. As illustrated in Table \ref{tab:ablation2}, we can see that when 1, 2, or 6 number of transformer heads are used, the network with 6 transformer heads performs better in all the experiments  than others for all the three evaluation protocols. This ablation shows that the number of transformer heads also effect the performance of the model and hence this factor is critical while designing the network for the SGG. This study motivated us to use six transformer heads in the novel BGT-Net.

\textbf{Performance with different number of Bidirectional GRU Layers.}
We compare networks with different number of BiGRU layers. To keep the comparison fair, we use all other same parameters in the experiments except the number of BiGRU layers. As shown in Table \ref{tab:ablation2}, it is evident that increasing the number of BiGRU layers does not significantly improve the performance, but it does increase the computational power and training time. Hence, in the BGT-Net, we only use one BiGRU layer.

\begin{table*}[!htb]
	\begin{minipage}{\textwidth}
		\centering
		\footnotesize
		\resizebox{\columnwidth}{!}{%
		\begin{tabular}{llllllccccccccccccccc}
		\hline
    	 &  \vline &\textbf{Transformer} & \textbf{GRU} & \textbf{FS} & \vline & \textbf{R @ 20} & \textbf{R @ 50} & \textbf{R @ 100} & \vline & \textbf{nG R @ 20} & \textbf{nG R @ 50} & \textbf{nG R @ 100} & \vline & \textbf{zs R @ 20} & \textbf{zs R @ 50} & \textbf{zs R @ 100} & \vline & \textbf{mR @ 20} & \textbf{mR @ 50} & \textbf{m R @ 100}\\
    	 \hline
	& \vline	& 	x & - & - & \vline & 23.61 & 30.4 &	34.81 & \vline & 24.23	& 32.88	& 39.06 & \vline & 0	& 0	& 0 & \vline & 4.62	& 6.55	& 7.85\\
		& \vline	&- & x & - & \vline & 24.1 &	31.2 &	35.5 & \vline & 25.37	& 34.59	& 41.14 & \vline & 0	& 0	& 0.03 & \vline& 4.07	& 5.49	& 6.51  \\
			SGDet& \vline&	- & - & x & \vline & 22.3 &	28.17 &	32.56 & \vline & 25.04	& 34.58	& 41.23 & \vline & 0.95	& 1.27	& 2.21 & \vline & 4.47	& 5.98	& 7.65\\
		& \vline	&x & x & x & \vline & 24.68 & 31.87 & 36.18 & \vline & 26.23	& 35.87	& 42.46 & \vline & 1.22	& 2.38	& 3.42 & \vline& 5.69	& 7.81	& 9.25\\
		\hline
		\hline
		& \vline &	x & - & - & \vline & 33.81	& 37.22	& 38.12 & \vline  & 38.55	& 46.28	& 50.04 & \vline & 0.15	& 0.45	& 0.7 & \vline & 7.31	& 9.14	& 9.71 \\
		& \vline&	- & x & - & \vline & 40.03	& 44	& 45.02 & \vline & 45.61	& 54.82	& 59.26 & \vline & 0.19	& 0.69	& 0.99 & \vline & 7.92	& 9.85	& 10.55 \\
		SGCls & \vline&	- & - & x & \vline & 35.63	& 38.92	& 39.77 & \vline  & 39.41	& 47.92	& 55.56 & \vline & 3.25	& 4.99	& 5.78 & \vline & 8.2	& 10.65	& 11.34\\
		& \vline&	x & x & x & \vline & 41.72	& 45.69	& 46.74 & \vline & 47.96	& 57.42	& 61.92 & \vline & 4.12	& 6.72	& 8.06 & \vline & 10.41	& 12.77	& 13.61\\
		\hline
		\hline
		 & \vline&   x & - & - & \vline & 57.98	& 64.75	& 66.63 & \vline & 65.92	& 80.51	& 87.82 & \vline & 0.64	& 2.06	& 3.69 & \vline & 12.14	& 15.59	& 17.05 \\
	 & \vline&		- & x & - & \vline & 58.52	& 65.19	& 67.08 & \vline & 65.76	& 80.38	& 87.83 & \vline & 0.68	& 2.52	& 4.46 & \vline & 12.33	& 15.79	& 17.16 \\
	PredCls	& \vline&	- & - & x & \vline & 56.73	& 63.48	& 65.53 & \vline  & 64.87	& 79.56	& 87.2 & \vline & 11.9	& 17.78	& 20.97 & \vline & 12.05	& 15.22	& 16.46\\
	& \vline &		x & x & x & \vline & 58.71	& 65.25	& 67.1 & \vline & 67.27	& 82.05	& 89.29 & \vline  & 12.06	& 18.22	& 21.49  & \vline & 14.36	& 17.88	& 19.44 \\
				\hline
		  	\end{tabular}}
		 \vspace{1mm}
  	\normalsize
	{\topcaption[Recall@K comparison of BGT-Net with Model Performance of different Modules.]{Ablation study performed on evaluation of three models. All the experimental factors are kept same except the ones which are being evaluated to have fair comparison. The evaluation is conducted using Recall@K, nGR@K, zsR@K, and mR@K. }	\label{tab:ablation1}} 
		\end{minipage} \hfill
		\vspace{-5mm}
\end{table*}

	\begin{table*}[!htb]
	\begin{minipage}{\textwidth}
		\centering
		\footnotesize
		\resizebox{\columnwidth}{!}{%
		\begin{tabular}{llllccccccccccccccc}
		\hline
    	 &  \vline &\textbf{Transformer Heads} &  \vline & \textbf{R @ 20} & \textbf{R @ 50} & \textbf{R @ 100} & \vline & \textbf{nG R @ 20} & \textbf{nG R @ 50} & \textbf{nG R @ 100} & \vline & \textbf{zs R @ 20} & \textbf{zs Recall @ 50} & \textbf{zs R @ 100} & \vline & \textbf{mR @ 20} & \textbf{mR @ 50} & \textbf{m R @ 100}\\
    	 \hline
	& \vline	& 1 & \vline & 23.78	& 31.15	& 35.4  & \vline &  25.76	& 34.8	& 41.78 & \vline & 0.99	& 2.11	& 2.67 & \vline	& 4.75	& 7.32	& 7.96\\
	SGDet	& \vline	& 2 & \vline  & 24.54	& 31.77	& 36.11 &  \vline & 26.05	& 35.64	& 42.39  & \vline & 1.07	& 2.18	& 3.37 & \vline & 5.57	& 7.52	& 8.8\\
			& \vline &	6 & \vline & \textbf{24.68}	& \textbf{31.87	}& \textbf{36.18} & \vline 	& \textbf{26.23}	& \textbf{35.87	}& \textbf{42.46} & \vline  & \textbf{1.22}	& \textbf{2.38}	& \textbf{3.42 }& \vline &\textbf{ 5.69}	& \textbf{7.81}	& \textbf{9.25}\\
		\hline
		\hline
	& \vline	& 1 & \vline & 38.62	& 42.54	& 43.66  & \vline & 45.01	& 52.1	& 58.78 & \vline & 2.67	& 4.31	& 4.89 & \vline 	& 8.21	& 9.98	& 10.32 \\
	SGCls	& \vline	& 2 & \vline  & 39.37	& 43.42	& 44.51&  \vline 	& 45.25	& 54.51	& 59.17  & \vline&  3.32	& 5.37	& 6.48 & \vline &  9.01	& 11.14	& 11.89 \\
	& \vline&	6 & \vline &\textbf{ 41.72
}	& \textbf{45.69}	& \textbf{46.74} & \vline 	& \textbf{47.96	}& \textbf{57.42}	& \textbf{61.92} & \vline   & \textbf{4.12}	& \textbf{6.72}	& \textbf{8.06} & \vline 		& \textbf{10.41}	& \textbf{12.77}	& \textbf{13.61 }\\	
	\hline
	\hline
	& \vline	& 1 & \vline &58.03	& 64.8	& 66.89 & \vline 	& 66.23	& 81.06	& 87.89  & \vline  & 11.06	& 17.39	& 19.45 & \vline & 12.55	& 16.21	& 17.51  \\
	PredCls	& \vline	& 2 & \vline & 58.45	& 64.98	& 67.03 &  \vline 	& 66.89	& 81.66	& 88.34 & \vline  & 11.9	& 18.11	& 21.34 & \vline & 12.91	& 16.75	& 18.92 \\
	& \vline&	6 & \vline & \textbf{58.71}	& \textbf{65.25}	& \textbf{67.1 }& \vline & \textbf{67.27}	& \textbf{82.05}	& \textbf{89.29} & \vline   & \textbf{12.06}	& \textbf{18.22}	&\textbf{ 21.49 }& \vline & \textbf{14.36}	& \textbf{17.88}	& \textbf{19.44} \\	
	\hline
  \end{tabular}}
  		 \vspace{1mm}
  	\normalsize
	{\topcaption[Model Performance for different Transformer Heads.]{Various R@K performance for the different numbers of Transformer Heads in BGT-Net using VG dataset.  }	\label{tab:ablation2}}
		\end{minipage} \hfill
	\vspace{-5mm}
\end{table*}

	\begin{table*}[!htb]
	\begin{minipage}{\textwidth}
		\centering
		\footnotesize
		\resizebox{\columnwidth}{!}{%
		\begin{tabular}{llllccccccccccccccc}
		\hline
    	 &  \vline &\textbf{Bi-GRU} &  \vline & \textbf{R @ 20} & \textbf{R @ 50} & \textbf{R @ 100} & \vline & \textbf{nG R @ 20} & \textbf{nG R @ 50} & \textbf{nG R @ 100} & \vline & \textbf{zs R @ 20} & \textbf{zs Recall @ 50} & \textbf{zs R @ 100} & \vline & \textbf{mR @ 20} & \textbf{mR @ 50} & \textbf{m R @ 100}\\
    	 \hline
	& \vline	& 1 & \vline& \textbf{25.54}	& \textbf{32.87}	& \textbf{37.3 }  & \vline & \textbf{27.24}	& \textbf{36.91}	& \textbf{43.72} & \vline & \textbf{4.8}	& \textbf{7.37}	& \textbf{8.78} & \vline	& \textbf{10.41	}& \textbf{12.77}	& \textbf{13.61}\\
	SGDet	& \vline	& 2 & \vline  & 24.54	& 31.77	& 36.11 &  \vline 	& 26.05	& 35.64	& 42.39  & \vline  & 1.07	& 2.18	& 3.37 & \vline & 9.91	& 12.28	& 13.12 \\
			& \vline &	6 & \vline & 23.93	& 31.01	& 35.37  & \vline 	& 25.47	& 35.02	& 41.63 & \vline  &  1.12	& 2.11	& 3.1 & \vline & 5.49	& 7.53	& 8.86 \\
		\hline
		\hline
	& \vline	& 1 & \vline & 41.69	& \textbf{45.96}	& \textbf{47.06} & \vline & 47.83 & \textbf{57.67}	& \textbf{62.29 }& \vline & 2.67	& 4.31	& 4.89 & \vline 	& 8.21	& 9.98	& 10.32 \\
	SGCls	& \vline	& 2 & \vline  & \textbf{41.72}	& 45.69	& 46.74 &  \vline & \textbf{47.96}	& 57.42	& 61.92  & \vline&\textbf{ 4.12}	& \textbf{6.72}	& \textbf{8.06} & \vline &  \textbf{9.01}	& \textbf{11.14}	& \textbf{11.89} \\
	
	& \vline&	6 & \vline &  41.08	& 45.57	& 46.23& \vline & 46.98	& 57.12	& 61.3 & \vline   & 4.02	& 6.56	& 7.79 & \vline & 8.54	& 11.81	& 13.01  \\	
	\hline
	\hline

	& \vline	& 1 & \vline &\textbf{ 59.21}	& \textbf{65.68}	& \textbf{67.45}  & \vline & \textbf{67.69}	&\textbf{ 82.42}	& \textbf{89.45}  & \vline  & \textbf{	12.23}	& \textbf{18.31}	& \textbf{21.51} & \vline & \textbf{14.7}	& \textbf{18.46}	& \textbf{20.08 } \\
	PredCls	& \vline	& 2 & \vline & 58.71	& 65.25	& 67.1 &  \vline & 67.27	& 82.05	& 89.29 & \vline  & 12.06	& 18.22	& 21.49	 & \vline &  14.36	& 17.88	& 19.44   \\
	& \vline&	6 & \vline & 58.22	& 65.45	& 66.74 & \vline &  66.91	& 81.87	& 88.69  & \vline   &  11.83	& 18.07	& 21.1 & \vline & 14.22	& 17.59	& 19.21\\	
	\hline

  \end{tabular}}
  		 \vspace{1mm}
  	\normalsize
	{\topcaption[Model Performance for different Numbers of Bidirectional GRU Layers.]{Various R@K performance comparison in relation to the number of BiGRU layers present in network when trained on VG dataset.}	\label{tab:ablation3}
}
		\end{minipage} \hfill
	\vspace{-5mm}
\end{table*} 

\begin{figure*}[!htb]
	\vspace{-2mm}
	\centering
	\includegraphics[width=1\linewidth, height=1.9cm]{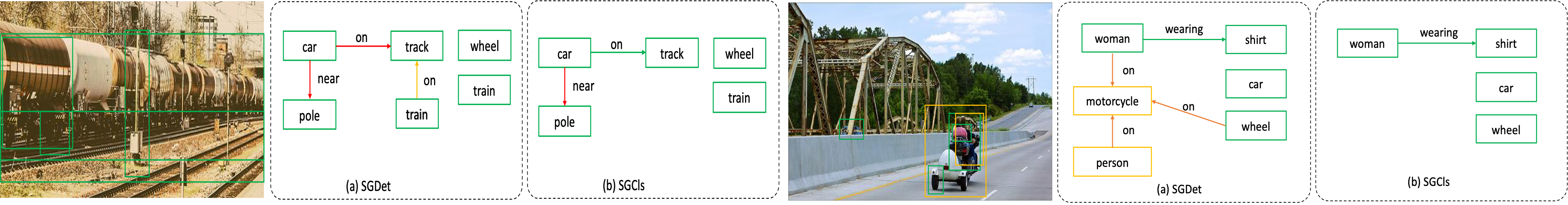}
    
	\vspace{-0.15cm}
	\caption{Qualitative results showing scene graphs generated by the BGT-Net. For both examples, (a) shows the scene graph generated in the SGDet protocol, and (b) the one generated in SGCls. The green arrow denotes that the detection object or relationship corresponding to the ground-truth. Orange arrows denote the detections that are not available in ground-truth but do represent the image properly. Red marks errors that are used for undetected relationships or wrongly detected objects.}
	\vspace{-6mm}
\label{fig:5}
\end{figure*}
\textbf{Qualitative Results.}
In Figure~\ref{fig:5}, Left: shows the qualitative results. In SGCls, the bounding boxes are given and the model has to predict the object class and the relationships. In SGDet, no information is given. In SGDet up to 160 objects in an image can be detected  but to keep the illustrations clean not every object detection is shown. It detected relationships present in the ground truth along with the additional feasible relationships. Figure~\ref{fig:5} Top: in SGCls the predicted scene graph fully corresponds to the ground truth scene graph for this image. There is no additional relationship predicted between other objects. Looking at the scene graph for SGDet, the difference between these two protocols can be seen very well. Also, the performance of the model is really good in this case. Additionally to the ground truth objects, the object ''motorcycle'' and ''person'' are detected. These two detections are correct and feasible. While the only ground truth relation (woman - wearing - shirt) is still being detected, three other relationships that are totally feasible are detected (woman - on - motorcycle), (person - on - motorcycle) and (wheel - on - motorcycle). The performance of the BGT-Net on this image is outstanding. No problems or specialties in the SGDet can be found.

In Figure~\ref{fig:5} Right: In SGDet, the model even fails to give the correct relationship (car - on - track). But, it correctly detects the whole train, which was not specified in the ground truth and the correct relationship (train - on - track). It is  special in this case and it might also be in a lot of other images that the model detects many objects that were not shown in the ground truth. It also shows a lot of relationships between these additionally shown objects. But, most likely with a higher amount of detected objects in an image, these triplets, that are not in the ground truth, leads the model to miss some of the relationships that would also be found in the ground truth.
\vspace{-3.5mm}
\section{Conclusion}
\vspace{-3mm}
We proposed a novel method BGT-Net to address the main challenges in SGG. BGT-Net solves the problems by 1) using the object-object communication by employing Bi-directional GRUs; 2) using transformer encoder with scaled-dot-product attention for predicting object classes after they have received feature information from other objects; 3) getting edge feature from second transformer encoder; 4) Utilising Frequency Softening and Bias Adaptation for dealing the with bias in the SGG. We validated the effectiveness of the proposed BGT-Net using extensive experiments and conducting elaborative ablation studies using three open-source datasets.
\enlargethispage{\baselineskip}

{\small
\bibliographystyle{ieee_fullname}
\bibliography{egbib}
}

\end{document}


\title{Supplementary Material for ``BGT-Net: Bidirectional GRU Transformer Network for Scene Graph Generation"}

\author{
 Naina Dhingra\textsuperscript{}\space\space\space\space  \hspace{40mm} Florian Ritter\textsuperscript{}\space\space\space\space  \hspace{40mm} Andreas
 Kunz\textsuperscript{}\space\space\space\space 
  \vspace{6px} \\
\textsuperscript{}Innovation Center Virtual Reality, ETH Zurich\space\space\space\space \\
{\tt\small $\{$ndhingra, kunz$\}$@iwf.mavt.ethz.ch, ritterf@ethz.ch}}

\maketitle

\begin{abstract}
The supplementary material is organized in the following manner: 1) section 1: a comprehensive review of BGT-Net without BiGRU layer; 2) section 2: more ablation study results; 3)  section 3: hyper-parameter study; 4) section 4: more qualitative results
\end{abstract}


\section{BGT-Net without Bi-directional GRU}

BGT-Net (no BiGRU) is the BGT-Net with no BiGRU in it. We experimented with it to see how Bi-directional GRU effects the performance of the network.

The performance of the BGT-Net (no BiGRU) for PredCls is lower than the other models. Even when compared to the MOTIFS, a performance decrease can be seen. In SGCls, there is slight improvements compared to the other models. Most important factor is that the performance increase compared to MOTIFS (baseline) is seen. Therefore, the effectiveness of the BGT-Net (no BiGRU) is shown in Table 5 in paper. 

In SGDet protocol, the BGT-Net (no BiGRU) can show an impressive performance. It outperforms every other model. The performance difference is also quite significant. The Recall@K is improved by over 1 point and reaches up to an increase of more than 3 points over the next best model. The results in mean Recall@K are worse than others. The short-comings of this BGT-Net (no BiGRU) is removed and improved by changing the model structure to BGT-Net by adding a Bi-GRU.

\section{Ablation Study}
As mentioned in the paper, we performed ablation study on several factors. We provide additional results on those experiments. They are discussed below: 

\subsection{Different Combination of Modules}
\label{subsubsec:model_building_blocks}
\begin{figure*}[!htb]
    \centering
    \includegraphics[width=\linewidth]{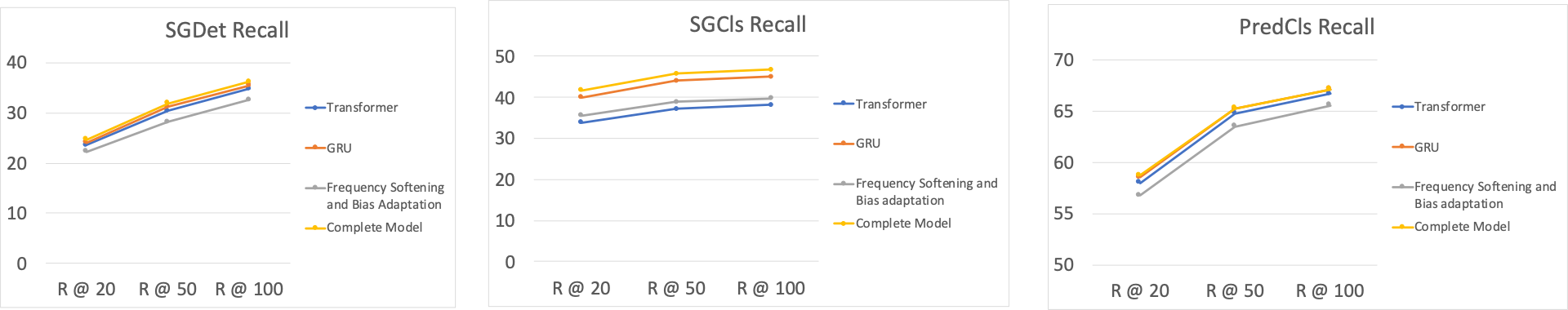} 
    \caption[Recall Results]{Graphical representation of Recall Results for SGDet (left), SGCls (middle), PredCls (right) comparing the effects of different modules of the BGT-Net made during ablation studies on the effectiveness of different modules.}
    \label{fig:abls1}
\end{figure*}

\begin{figure*}[!htb]
    \centering
    \includegraphics[width=\linewidth]{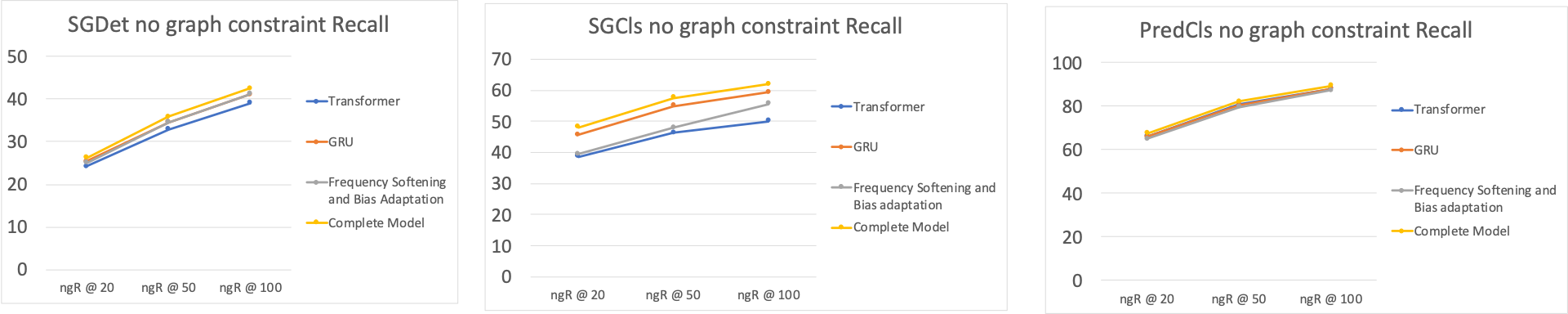} 
    \caption[No Graph Constraint Recall Results]{Graphical representation of no graph constraint Recall Results for SGDet (left), SGCls (middle), PredCls (right) comparing the effects of different modules of the BGT-Net made during ablation studies on the effectiveness of different modules.}
    \label{fig:abls2}
\end{figure*}

\begin{figure*}[!htb]
    \centering
    \includegraphics[width=\linewidth]{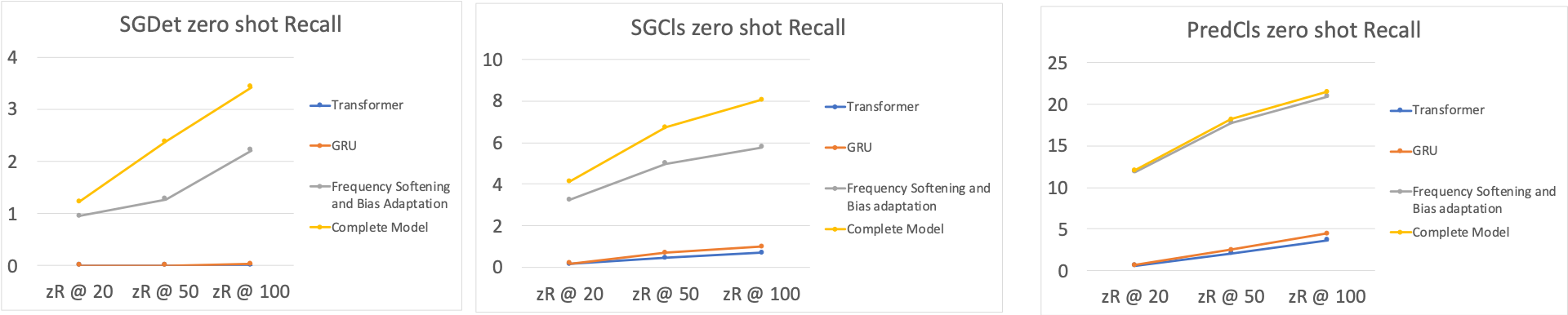} 
    \caption[Zero Shot Recall Results]{Graphical representation of zero shot Recall Results for SGDet (left), SGCls (middle), PredCls (right) comparing the effects of different modules of the BGT-Net made during ablation studies on the effectiveness of different modules.}
    \label{fig:abls3}
\end{figure*}

\begin{figure*}[!htb]
    \centering
    \includegraphics[width=\linewidth]{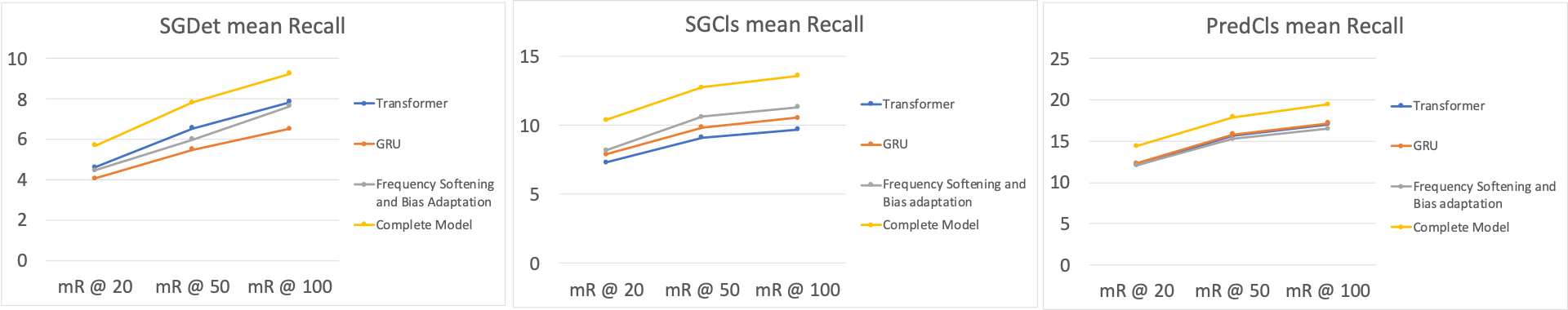} 
    \caption[Mean Recall Results]{Graphical representation of mean Recall Results for SGDet (left), SGCls (middle), PredCls (right) comparing the effects of different modules of the BGT-Net made during ablation studies on the effectiveness of different modules.}
    \label{fig:abls4}
\end{figure*}

Figure \ref{fig:abls1}, Figure \ref{fig:abls2}, Figure \ref{fig:abls3}, and Figure \ref{fig:abls4} illustrate the graphical representation of the performance achieved using different modules by evaluating them on different performance recall metric.

\subsection{Number of Transformer Heads}
\begin{figure*}[!htb]
    \centering
    \includegraphics[width=\linewidth]{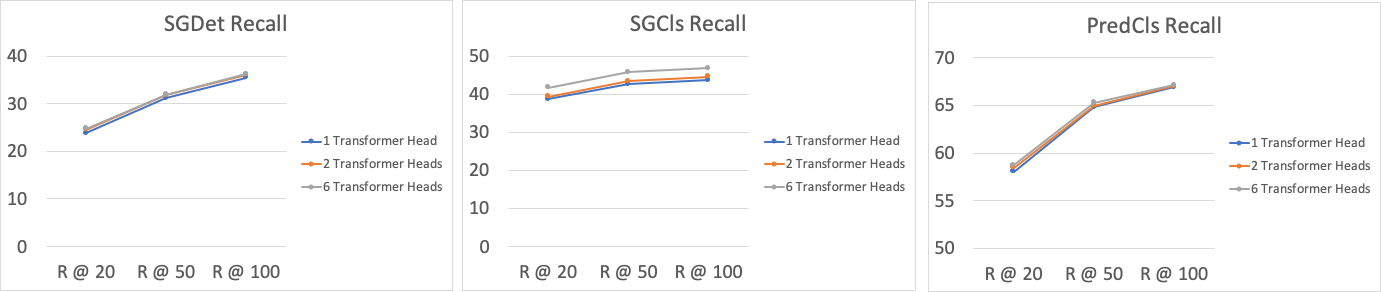} 
    \caption[Recall Results]{Graphical representation of Recall Results for SGDet (left), SGCls (middle), PredCls (right). Evaluating the performance changes evoked by changing the number of Transformer heads of Transformer Encoders for object and edge information.}
    \label{fig:abls-th1}
\end{figure*}

\begin{figure*}[!htb]
    \centering
    \includegraphics[width=\linewidth]{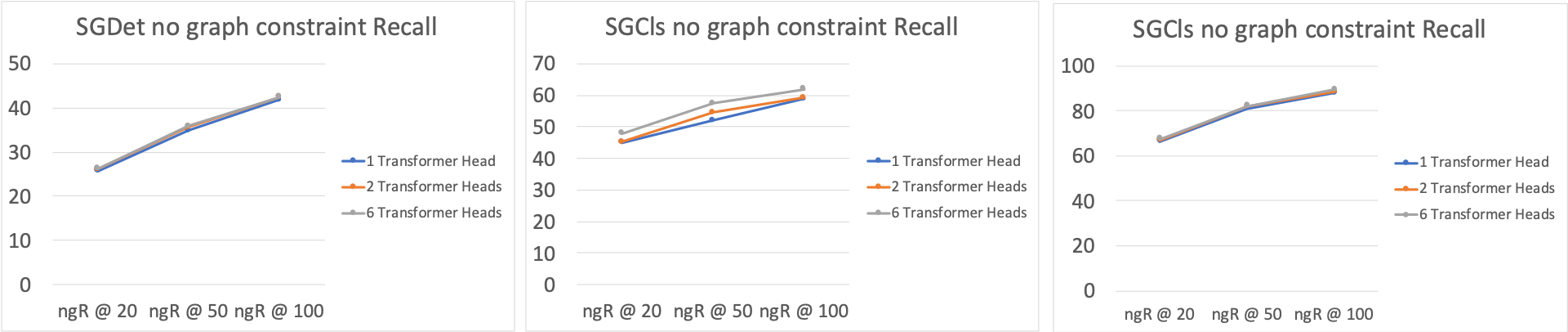} 
    \caption[No Graph Constraint Recall Results]{Graphical representation of no graph constraint Recall Results for SGDet (left), SGCls (middle), PredCls (right). Evaluating the performance changes evoked by changing the number of Transformer heads of Transformer Encoders for object and edge information.}
    \label{fig:abls-th2}
\end{figure*}

\begin{figure*}[!htb]
    \centering
    \includegraphics[width=\linewidth]{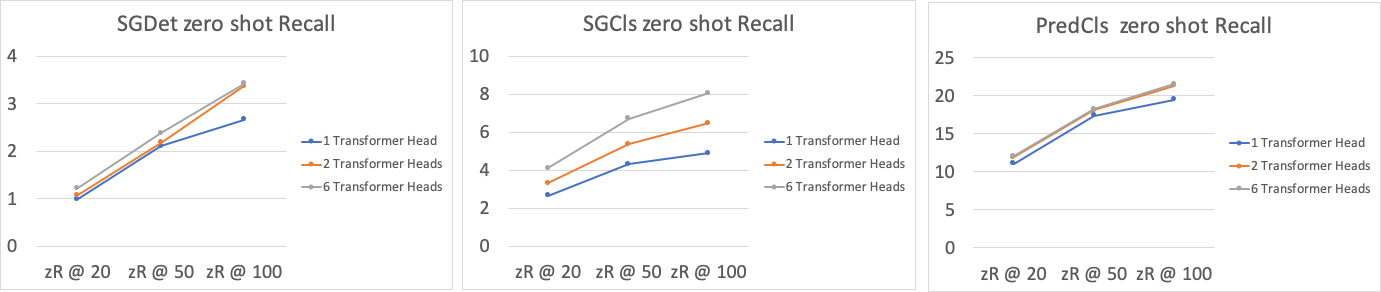} 
    \caption[Zero Shot Recall Results]{Graphical representation of zero shot Recall Results for SGDet (left), SGCls (middle), PredCls (right). Evaluating the performance change evoked by changing the number of Transformer heads of Transformer Encoders for object and edge information.}
    \label{fig:abls-th3}
\end{figure*}

\begin{figure*}[!htb]
    \centering
    \includegraphics[width=\linewidth]{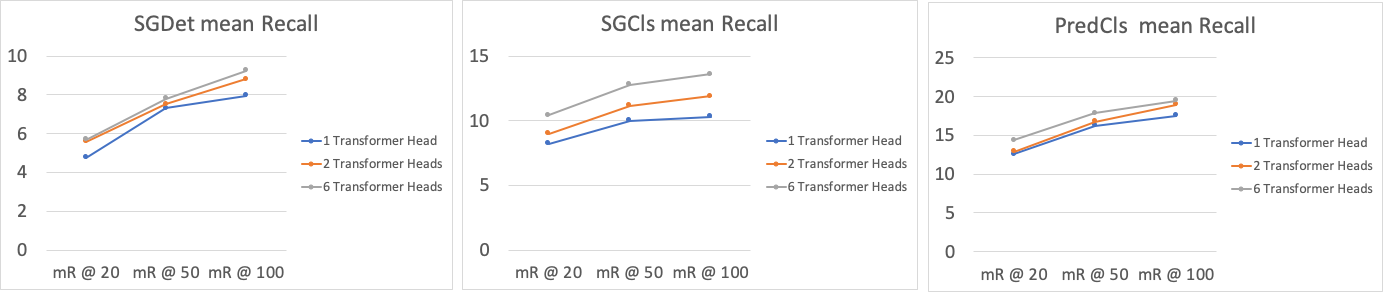} 
    \caption[Mean Recall Results]{Graphical representation of mean Recall Results for SGDet (left), SGCls (middle), PredCls (right). Evaluating the performance change evoked by changing the number of Transformer heads of Transformer Encoders for object and edge information.}
    \label{fig:abls-th4}
\end{figure*}

Figure \ref{fig:abls-th1}, Figure \ref{fig:abls-th2}, Figure \ref{fig:abls-th3}, and Figure \ref{fig:abls-th4} show the graphical representation of the performance achieved by using different number of transformer heads and by varying the performance recall metric.

\subsection{Number of Bidirectional GRU Layers}
\begin{figure*}[!htb]
    \centering
    \includegraphics[width=\linewidth]{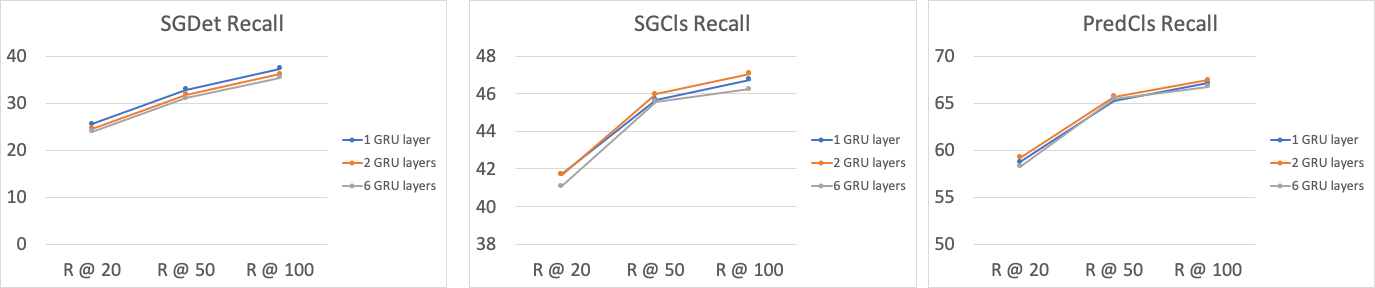} 
    \caption[Recall Results]{Graphical representation of Recall Results for SGDet (left), SGCls (middle), PredCls (right). Evaluation of performance change influenced by using 1, 2 or 6 layers of bidirectional GRUs in the BGT-Net model.}
    \label{fig:abls-gru1}
\end{figure*}

\begin{figure*}[!htb]
    \centering
    \includegraphics[width=\linewidth]{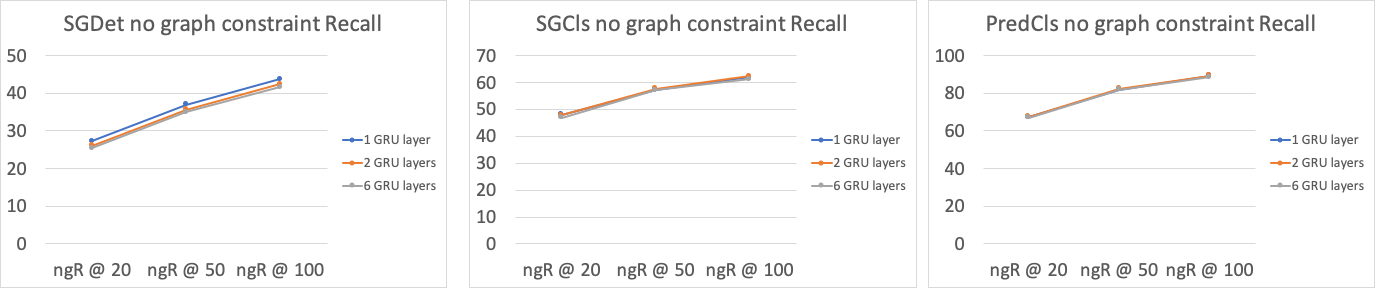} 
    \caption[No Graph Constraint Recall Results]{Graphical representation of no graph constraint Recall Results for SGDet (left), SGCls (middle), PredCls (right). Evaluation of performance change influenced by using 1, 2, or 6 layers of bidirectional GRUs in the BGT-Net model.}
    \label{fig:abls-gru2}
\end{figure*}

\begin{figure*}[!htb]
    \centering
    \includegraphics[width=\linewidth]{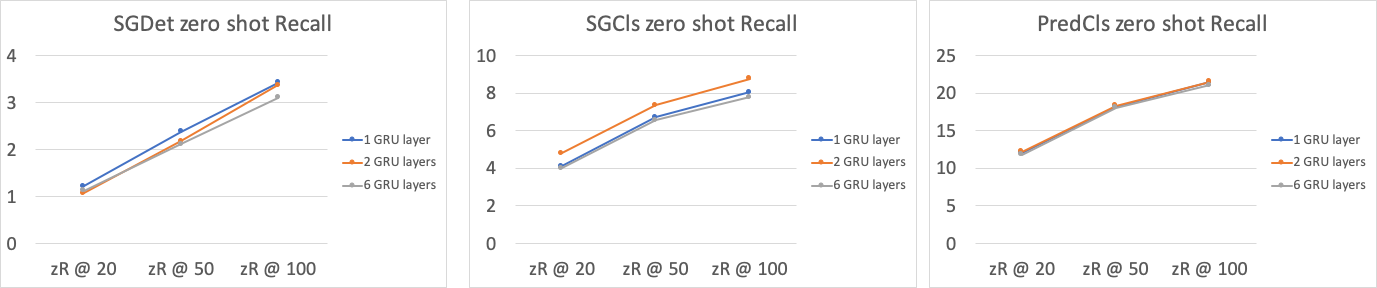} 
    \caption[Zero Shot Recall Results]{Graphical representation of zero shot Recall Results for SGDet (left), SGCls (middle), PredCls (right). Evaluation of performance change influenced by using 1, 2, or 6 layers of bidirectional GRUs in the BGT-Net model.}
    \label{fig:abls-gru3}
\end{figure*}

\begin{figure*}[!htb]
    \centering
    \includegraphics[width=\linewidth]{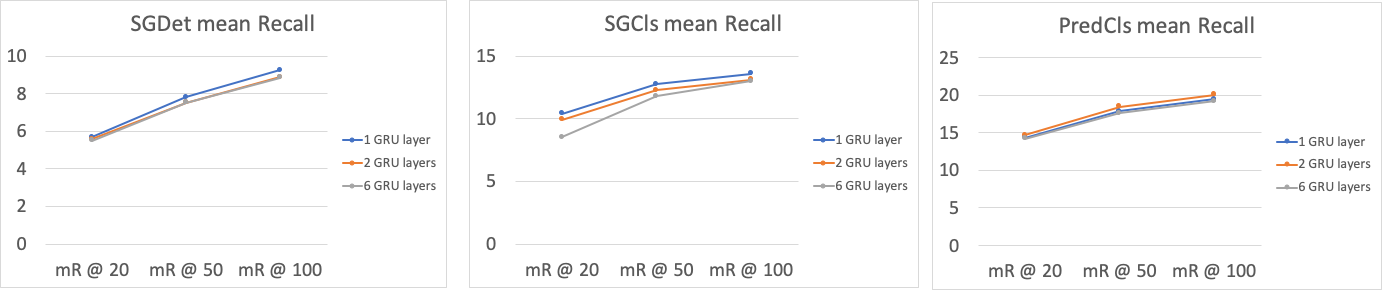} 
    \caption[Mean Recall Results]{Graphical representation of mean Recall Results for SGDet (left), SGCls (middle), PredCls (right). Evaluation of performance change influenced by using 1, 2, or 6 layers of bidirectional GRUs in the BGT-Net model.}
    \label{fig:abls-gru4}
\end{figure*}

 Figure \ref{fig:abls-gru1}, Figure \ref{fig:abls-gru2}, Figure \ref{fig:abls-gru3}, and Figure \ref{fig:abls-gru4} show the graphical representation of the performance achieved by using different number of BiGRU layers and by varying the performance recall metric.
 
\section{Hyper-parameter Study}

The influence of the most important hyper-parameters on the model performance were tested. Batch size, learning rate, and number of solver iterations were varied for multiple experiments. 

PredCls was the main protocol on which the performance was compared. One hyper-parameter at a time was varied to understand its influence. The effect of changing parameters and therefore showing which parameter set performs the best can be seen below.

\textbf{Leraning Rate.} Batch size was fixed at 24 and number of solver iterations at 24000. Learning rates 0.0001, 0.0005, 0.001 and 0.002 were tested. 

\begin{figure}[!htb]
    \centering
    \includegraphics[width=\linewidth]{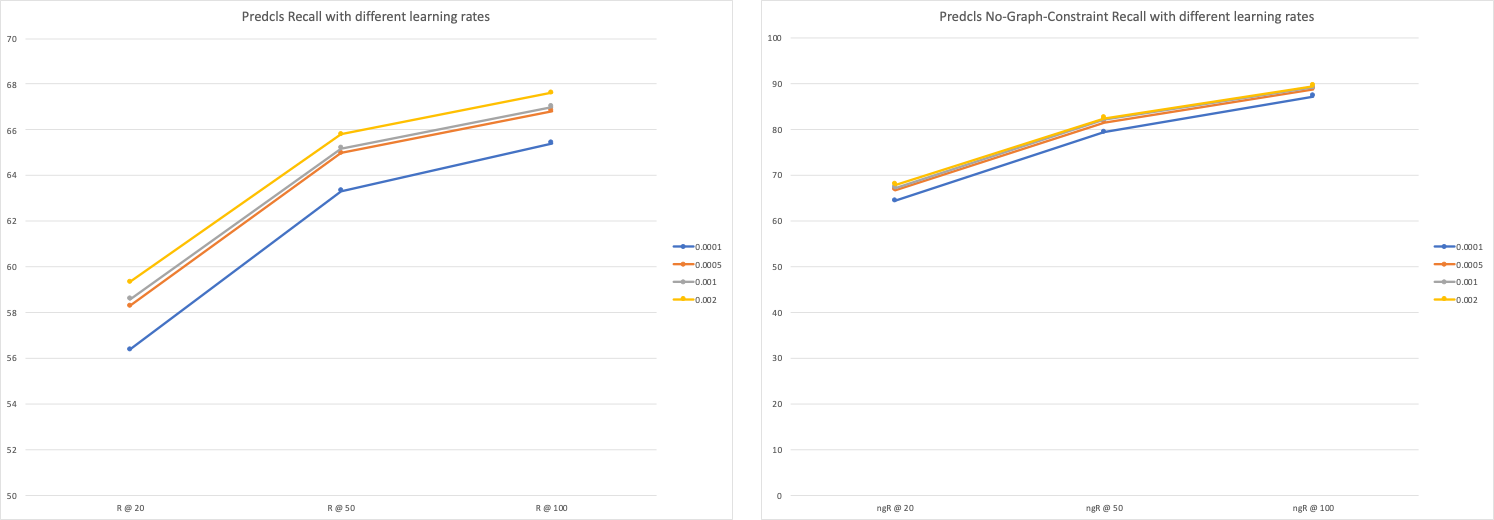} 
    \caption[PredCls Results for different Learning Rates 1]{Recall@K (left) and no graph constraint Recall@K (right) for Predicate Classification using different learning rates}
    \label{fig:gru_tf-hyp1}
\end{figure}

As \figref{fig:gru_tf-hyp1}, shows the best performance learning rate when evaluated using Recall@K. Evaluating on no graph constraint Recall@K illustrates no significant difference by the influence of learning rate.

\begin{figure}[!htb]
    \centering
    \includegraphics[width=\linewidth]{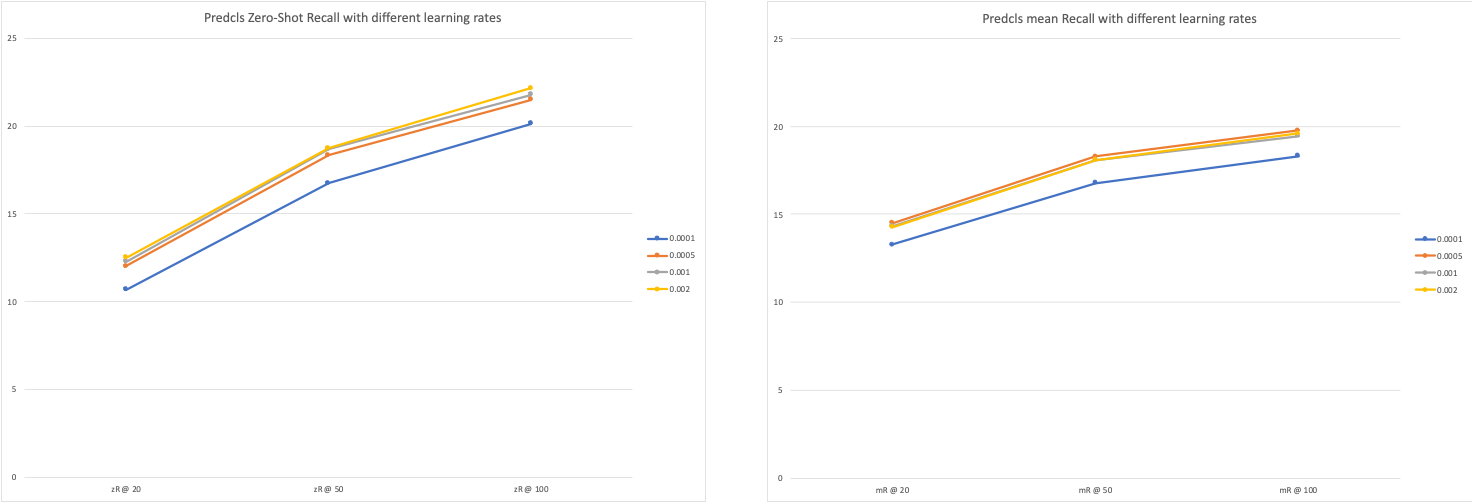} 
    \caption[PredCls Results for different Learning Rates 2]{Zero shot Recall@K (left) and mean Recall@K (right) for Predicate Classification using different learning rates}
    \label{fig:gru_tf-hyp2}
\end{figure}

Similarly, in \figref{fig:gru_tf-hyp2}, the influence of the learning rate for evaluation on zero shot and mean Recall@K can not clearly be seen. Performance of learning rates 0.002, 0.001 and 0.00005 are almost the same. But learning rate 0.002 seems to have a slight edge on the other two. But this difference is marginal. Only the smallest learning rate 0.0001 seems to be under-performing.

\newpage

\textbf{Batch Size.} The non-changed hyper-parameters are set to 0.002 for the learning rate and 24000 to solver iterations. The examined batch sizes are 6, 12, 18 and 24. 

\begin{figure}[!htb]
    \centering
    \includegraphics[width=\linewidth]{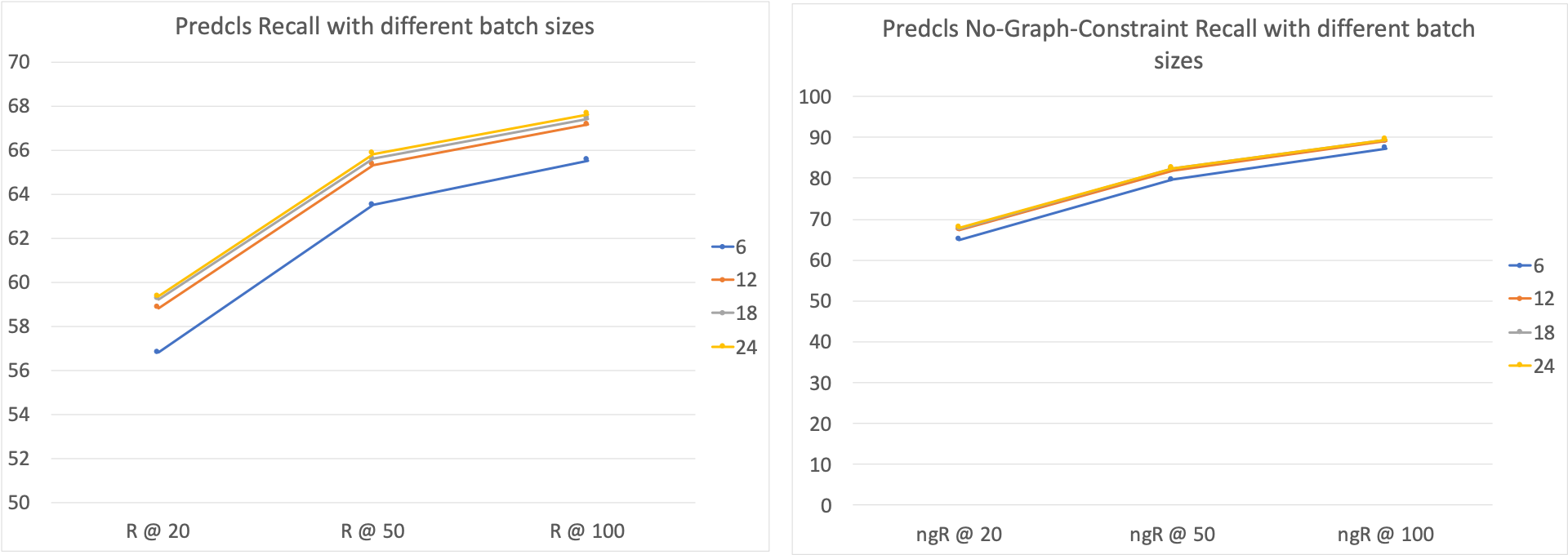} 
    \caption[PredCls Results for different Batch Sizes 1]{Recall@K (left) and no graph constraint Recall@K (right) for Predicate Classification using different batch sizes}
    \label{fig:gru_tf-hyp3}
\end{figure}

\begin{figure}[!htb]
    \centering
    \includegraphics[width=\linewidth]{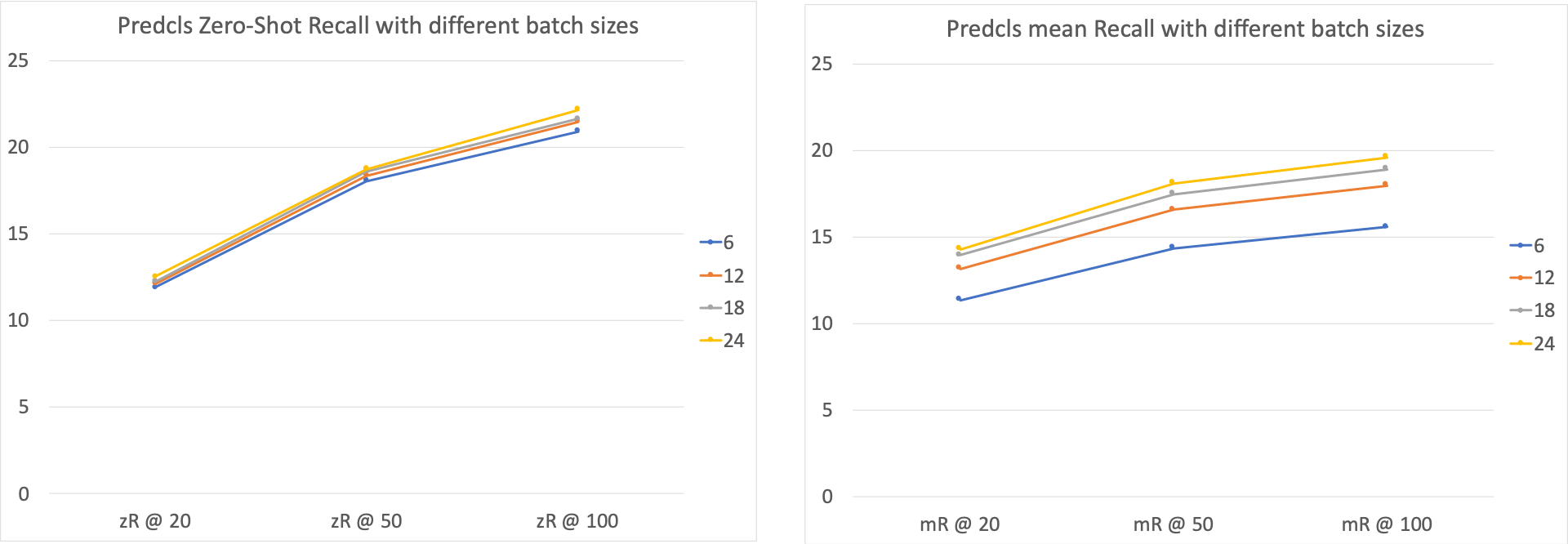} 
    \caption[PredCls Results for different Batch Sizes 2]{Zero shot Recall@K (left) and mean Recall@K (right) for Predicate Classification using different batch sizes}
    \label{fig:gru_tf-hyp4}
\end{figure}

Only the smallest batch size 6 shows the lower performance in Recall@K as it can be seen in \figref{fig:gru_tf-hyp3}. This difference disappears for the no graph constraint Recall@K (visible in \figref{fig:gru_tf-hyp3} on the right) and for the zero shot Recall@K (shown in \figref{fig:gru_tf-hyp4} on the left). 

Significant and most evident difference in performance can be seen in the mean Recall@K in \figref{fig:gru_tf-hyp4}. Clearly, the largest batch size is performing better in this metric. Also in the other metrics, batch size 24 has the highest values.

\newpage

\textbf{Solver Iterations.} Keeping the learning rate at 0.002 and the batch size at 24, while changing the solver iterations to 6000, 12000, 18000 and 24000 shows the following influence of the solver iterations on the model performance.

\begin{figure}[!htb]
    \centering
    \includegraphics[width=\linewidth]{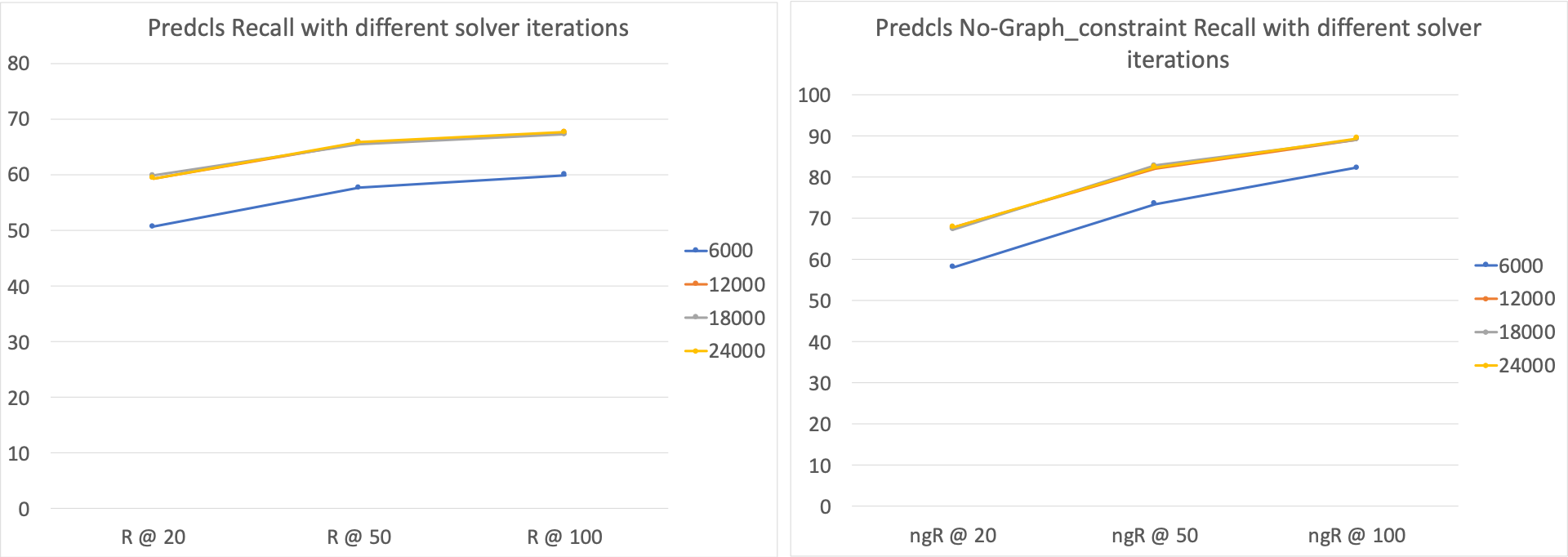} 
    \caption[PredCls Results for different Solver Iterations 1]{Recall@K (left) and no graph constraint Recall@K (right)  for Predicate Classification compared to different solver iterations}
    \label{fig:gru_tf-hyp5}
\end{figure}

\begin{figure}[!htb]
    \centering
    \includegraphics[width=\linewidth]{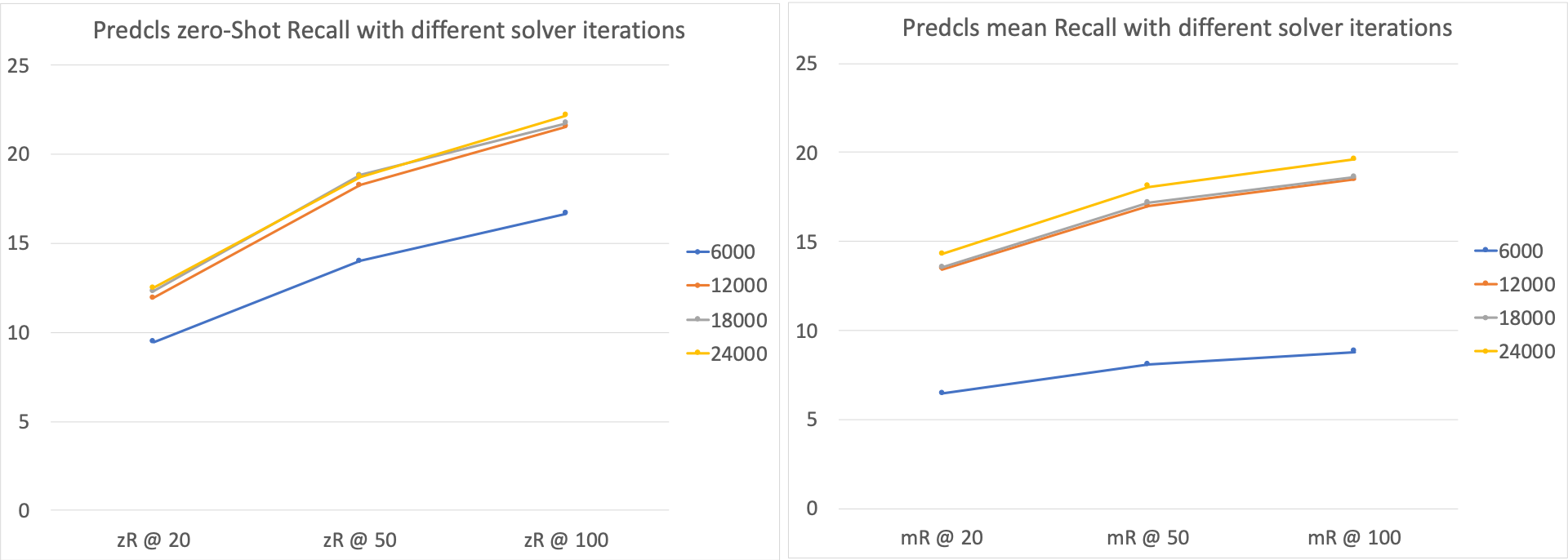} 
    \caption[PredCls Results for different Solver Iterations 2]{Zero shot Recall@K (left) and mean Recall@K (right)  for Predicate Classification compared to different solver iterations}
    \label{fig:gru_tf-hyp6}
\end{figure}

Throughout all the results in \figref{fig:gru_tf-hyp5} and \figref{fig:gru_tf-hyp6}, 6000 solver iterations under-perform significantly. Having only 6000 iterations does not allow the model to converge. As before with batch size, the difference in performance while changing the solver iterations is really small. Only for mean Recall@K the highest solver iteration does improve the results by almost 1 point. 

After these results, the best performing set of hyper-parameters can be found. Combining the results for batch size, learning rate and solver iterations leads to the choice of a learning rate of 0.002 a batch size of 24 and a solver iteration of 24000. Possibly, the least important of these would be the solver iterations since only the smallest difference going from 18000 to 24000 was detected.

\section{Qualitative Results: BGT-Net}

\begin{figure*}[!htb]
	\centering
	\includegraphics[width=1\linewidth]{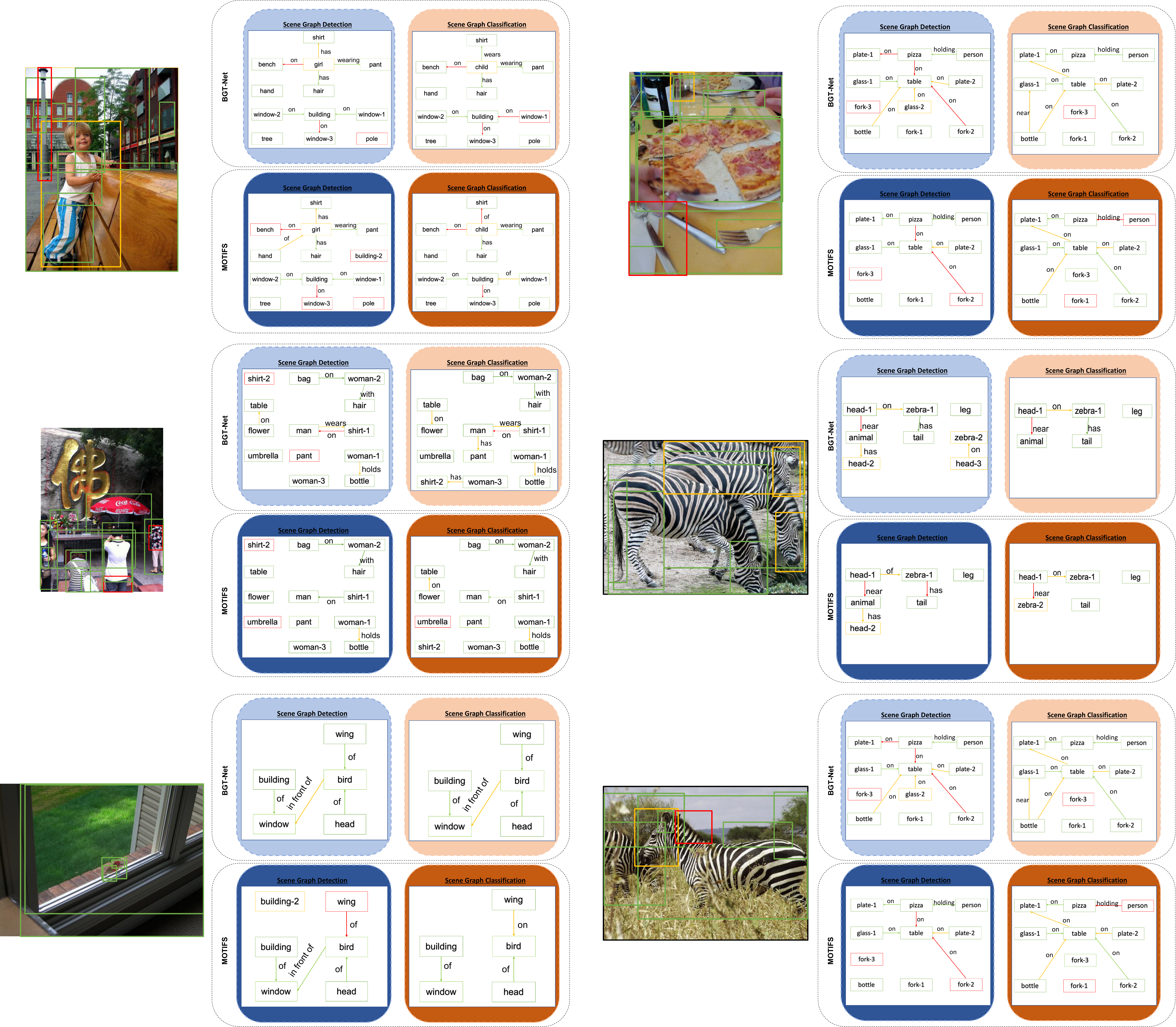}
    
	\vspace{-0.15cm}
	\caption{Qualitative results of BGT-Model generated scene graphs. Two protocols are shown. left: Scene Graph Detection (SGDet), right: Scene Graph Classification (SGCls). BGT-Net is qualitatively compared to the MOTIFS model \cite{zellers2018neural}. Three colours are used to specify properties of detections. `Green' show detections that also perfectly correspond with ground-truth. `Red' is used for wrong detections and `orange' for detections not available in ground-truth but when checking with visual scene still represent the situation correctly.}
	\vspace{-3mm}
\label{fig:qual_res}
\end{figure*}

The qualitative results in Figure \ref{fig:qual_res} show generated scene graphs on images of the Visaul Genome dataset. The examples are compared to the scene graphs generated by the MOTIFS model \cite{zellers2018neural}. For illustration purpose, the objects and relationships are coloured to represent properties of these detection. Object or relationship coloured ”green” are detected properly and correspond to the ground-truth. `Red' shows wrongly detected entities. For objects this can be either due to incorrect class prediction or due to not detection of the object during detection step with the Faster R-CNN. `Orange' denotes detections which do not correspond with the ground-truth annotations but can be validated by human inspection which means that the prediction generally corresponds with the visual scene.

In Figure \ref{fig:qual_res}, the scene graphs generated with the Scene Graph Detection (SGDet) protocol are shown on the left and the ones generated with the Scene Graph Classification (SGCls) protocol are shown on the right. In SGDet, in many images, a large amount of objects are being detected. To increase readability, the object connected to the ground-truth objects are additionally inserted into the scene graph (with the colour `orange'). In many images, lots of $<$ subject- object-relationship$>$ triplets get correctly predicted but these cannot be found in the ground-truth triplets. This can effect the the performance of the model since these maybe correctly detected but not annotated in ground-truth and influence the prediction of other relationships. 

Both of the compared models in Figure \ref{fig:qual_res}, show errors in their predictions. But while detecting an object wrongly happens quite rarely, the relationship prediction is still much more prone to errors. This can be just the problem of the model but it is much more likely that many of the predictions made are not essentially wrong. This can be illustrated with the example of the object pair `person' and `pants'. Most often a person `wears' their pants. Relationships like `has', `in' and `on' do not contradict the reality. Added difficulty lays in the fact that throughout the dataset, these mentioned relationships do recently also appear. But there cannot be an evidence in the image what the relationship in this case can be because the difference between $<$person-pants-wears$>$ and $<$person-pants-has$>$ will not be visible in the image. This leads to worse model performance directly induced by the characteristics of the dataset. 
 
As shown in \ref{fig:qual_res}, errors in object prediction in SGCls do not happen very often. This may be the result of the highly improved performance of the BGT-Net in this protocol when measured in all evaluated metrics. Compared to the other protocols, the gained performance in SGCls is proportionally higher.

{\small
\bibliographystyle{ieee_fullname}
\bibliography{egbib}
}